\documentclass[11pt]{article}
\usepackage{fullpage}
\usepackage{titletoc}
\usepackage[page, header, toc, page]{appendix}
\usepackage[utf8]{inputenc} % allow utf-8 input
\usepackage[T1]{fontenc}    % use 8-bit T1 fonts
\usepackage{hyperref}       % hyperlinks
\usepackage{url}            % simple URL typesetting
\usepackage{booktabs}       % professional-quality tables
\usepackage{amsfonts}       % blackboard math symbols
\usepackage{nicefrac}       % compact symbols for 1/2, etc.
\usepackage{microtype}      % microtypography
\usepackage{tcolorbox}
\usepackage{diagbox}
\usepackage{enumerate}
\usepackage[shortlabels]{enumitem}
\usepackage[vlined, linesnumbered, ruled]{algorithm2e}
\usepackage{subfigure}
\usepackage{graphicx} % more modern
\usepackage{caption}
\usepackage{amsmath}
\usepackage{amsthm}
\usepackage{amssymb}
\usepackage{tikz}
\usepackage{tablefootnote}
\usepackage{multirow}
\usepackage{enumerate}
\usepackage{color}
\usepackage{xcolor}
\usepackage{natbib}

\allowdisplaybreaks[4]

\usepackage{mathrsfs}

\usepackage{hyperref}
\usepackage{bm,bbm,todonotes}

\allowdisplaybreaks

\newtheorem{Rema}{Remark}[section]
\newcommand{\eq}[1]{Eq.~\eqref{#1}}

\def\hyz#1 {\textcolor{red}{Hyz: #1 }}
\def\lx#1 {\textcolor{magenta}{Lx: #1 }}

\hypersetup{
	colorlinks=true,
	filecolor=blue,
	citecolor = blue,
	urlcolor=cyan,
}

\newcommand{\widesim}[2][1.5]{\mathrel{\overset{#2}{\scalebox{#1}[1]{$\sim$}}}}
\newcommand{\iidsim}{\widesim[2.33]{\mathrm{i.i.d.}}}	% independent and identically distributed
\newcommand{\N}{\mathcal{N}}				% normal
\newcommand{\1}{\mathbbm{1}}				% indicator
\renewcommand{\P}{\mathbb{P}}				% probability
\newcommand{\E}{\mathbb{E}}					% expectation
\newcommand{\agp}[1]{\langle #1 \rangle}	% angle process
\newcommand{\normmm}[1]{\lvert\kern-0.25ex\lvert\kern-0.25ex\lvert #1 \rvert\kern-0.25ex\rvert\kern-0.25ex\rvert}

\newcommand{\R}{\mathbb{R}}		% reals
\renewcommand*{\d}{\mathop{}\!\mathrm{d}}	% differential
\newcommand{\pd}{\partial}					% partial differential
\newcommand{\e}{\mathrm{e}}

\newcommand{\argmin}{\operatornamewithlimits{arg\,min}}
\newcommand{\argmax}{\operatornamewithlimits{arg\,max}}
\newtheorem{Def}{Definition}[section]
\newtheorem{Lem}[Def]{Lemma}
\newtheorem{Thm}[Def]{Theorem}

\title{
	Follow-the-Perturbed-Leader Approaches Best-of-Both-Worlds for the $m$-Set Semi-Bandit Problems
}

\author{
Jingxin Zhan 
\thanks{School of Mathematical Sciences, Peking University; email: \texttt{bjdxzjx@pku.edu.cn}.}
\and
Yuchen Xin 
\thanks{School of Mathematical Sciences, Peking University; email: \texttt{2301110087@pku.edu.cn}.}
\and
Chenjie Sun
\thanks{School of Mathematical Sciences, Peking University; email: 
\texttt{scj233@stu.pku.edu.cn}.}
\and
Zhihua Zhang 
\thanks{School of Mathematical Sciences, Peking University; email: \texttt{zhzhang@math.pku.edu.cn}. }
}

\begin{document}

\maketitle

\begin{abstract}
  We consider a common case of the combinatorial semi-bandit problem, the $m$-set semi-bandit, where the learner exactly selects $m$ arms from the total $d$ arms. In the adversarial setting, the best regret bound, known to be $\mathcal{O}(\sqrt{nmd})$ for time horizon $n$, is achieved by the well-known Follow-the-Regularized-Leader (FTRL) policy. However, this requires to explicitly compute the arm-selection probabilities via optimizing problems at each time step and sample according to them. This problem can be avoided by the Follow-the-Perturbed-Leader (FTPL) policy, which simply pulls the $m$ arms that rank among the $m$ smallest (estimated) loss with random perturbation. In this paper, we show that FTPL with a Fr\'echet perturbation also enjoys the near optimal regret bound $\mathcal{O}(\sqrt{nm}(\sqrt{d\log(d)}+m^{5/6}))$ in the adversarial setting and approaches best-of-both-world regret bounds, i.e., achieves a logarithmic regret for the stochastic setting. Moreover, our lower bounds show that the extra factors are unavoidable with our approach; any improvement would require a fundamentally different and more challenging method.
\end{abstract}

\section{Introduction}
The combinatorial semi-bandit problem
\citep{CESABIANCHI20121404} is an important online decision-making problem with partial information feedback, and has many practical applications such as in shortest-path problems \citep{gai2012combinatorial}, ranking \citep{pmlr-v37-kveton15}, multi-task bandits \citep{CESABIANCHI20121404} and recommender systems \citep{zou2019reinforcement}. The semi-bandit problem is a sequential game that involves a learner and an environment, both interacting over time. In particular, 
the problem setup consists of \( d \) fixed arms, and at each round \( t = 1, 2, \ldots \), the learner selects a combinatorial action—a subset of arms—from a predefined set \( \mathcal{A} \subset \{0,1\}^d \). Simultaneously, the environment generates a loss vector \( \ell_t \in [0,1]^d \). The learner then incurs a loss of \( \langle A_t, \ell_t \rangle \), where \( A_t \in \mathcal{A} \) is the selected action, and receives semi-bandit feedback \( o_t = A_t \odot \ell_t \), representing the losses associated with the selected arms only (here, \( \odot \) denotes element-wise multiplication). 

In this work, we focus on a common instance of the semi-bandit setting, the $m$-set semi-bandit \citep{Matroidbandits}, where each action consists of exactly \( m \) arms. That is, the action set is given by  
$
\mathcal{A} = \{ a \in \{0,1\}^d \,:\, \|a\|_1 = m \},
$  
with \( 1 \leq m \leq d \). 
The performance of the learner is quantified by the pseudo-regret, defined as  
$
\operatorname{Reg}_n := \mathbb{E}\left[\sum_{t=1}^n \left\langle A_t - a_\star, \ell_t \right\rangle\right],
$  
where \( a_\star = \arg\min_{a \in \mathcal{A}} \mathbb{E}\left[\sum_{t=1}^n \langle a, \ell_t \rangle\right] \) represents the optimal fixed action in hindsight. The expectation is taken over the randomness of both the learner’s decisions and the loss. The combinatorial semi-bandit problem has been studied primarily under two frameworks: the stochastic setting and the adversarial setting.

In the adversarial setting, no assumptions are made about the generation of the loss vectors \( \ell_t \); they can be chosen arbitrarily, possibly in an adaptive manner \citep{pmlr-v37-kveton15,pmlr-v40-Neu15,pmlr-v80-wang18a}. The optimal regret bound is $\mathcal{O}(\sqrt{nmd})$ \citep{audibert2014regret} (when $m\leq d/2$).  
In the stochastic setting, the losses \( \ell_1, \ell_2, \ldots, \ell_n \in [0,1]^d \) are independent and identically distributed samples drawn from an unknown but fixed distribution \( \mathcal{D} \). For each arm \( i \in \{1, \dots, d\} \), the expected loss is denoted by \( \nu_i = \mathbb{E}_{\ell \sim \mathcal{D}}[\ell_i] \in [0,1] \). The suboptimality gap of arm $i$ is expressed by $\Delta_i:=(\nu_i-\max\limits_{\mathcal{I}\subset\{1,\cdots,d\},|\mathcal{I}|<m}\min\limits_{j\notin \mathcal{I}}\nu_j
 )^+$ and the minimum gap is $\Delta=\min\limits_{1\leq i\leq d, \; 0< \Delta_i}\Delta_i$. There are many algorithms that were shown to achieve logarithmic regrets. For example, \cite{pmlr-v37-kveton15} and \cite{pmlr-v80-wang18a} derived $\mathcal{O}(\frac{(d-m)\log(n)}{\Delta})$ regrets in $m$-set semi-bandits.

In real-world scenarios, it is often unclear whether the environment follows a stochastic or adversarial pattern, making it desirable to design policies that offer regret guarantees in both settings. To address this challenge, particularly in the classical multi-armed bandit setting, a line of research has focused on Best-of-Both-Worlds (BOBW) algorithms, which aim to achieve near-optimal performance in both regimes. A pioneering contribution in this direction was made by \cite{pmlr-v23-bubeck12b}, who introduced the first BOBW algorithm. More recently, the well-known Tsallis-INF algorithm was proposed by \cite{pmlr-v89-zimmert19a}. In the context of combinatorial semi-bandits, related advancements have been made by \cite{zimmert2019beating}, \cite{NEURIPS2021_15a50c8b} and \cite{pmlr-v206-tsuchiya23a}.

However, most existing BOBW algorithms are 
Follow-the-Regularized-Leader (FTRL) policies and require to explicitly compute the arm-selection probabilities by solving optimizing problems at each time step and sample according to it. This problem, particularly in combinatorial semi-bandits \citep{pmlr-v40-Neu15}, has attracted interest and can be avoided by the Follow-the-Perturbed-Leader (FTPL) policy, which simply pulls the $m$ arms that rank among the $m$ smallest (estimated) loss with random perturbation. More precisely, the FTPL algorithm selects the action  
$
\arg\min_{a \in \mathcal{A}} \big\langle \hat{L}_t - \frac{r_t}{\eta_t},\, a \big\rangle,
$
where \( r_{t,i} \) denotes a random perturbation drawn from a specified distribution, \( \eta_t \) is the learning rate, and \( \hat{L}_{t,i} \) is an estimate of the cumulative loss for arm \( i \), defined as \( L_{t,i} = \sum_{s=1}^{t-1} \ell_{s,i} \).

\cite{pmlr-v201-honda23a} first proved that FTPL with Fr\'echet perturbations of shape parameter \( \alpha = 2 \) successfully achieves BOBW guarantees in the original bandit setting (i.e., when \( m = 1 \)), which was recently generalized by \cite{pmlr-v247-lee24a}. They analyzed general Fr\'echet-type tail distributions and underscored the effectiveness of the FTPL approach. Nevertheless, in  $m$-set semi-bandits the arm-selection probabilities $w_{t,i}=\phi_i(\eta_t \hat{L}_t)$ are much more complicated compared to the original setting,  making it harder to analyze the regret for FTPL.

\subsection{Contribution}
In this work, we show that FTPL with Fr\'echet perturbations achieves $\mathcal{O}(\sqrt{nm}(\sqrt{d\log(d)}+m^{5/6}))$ regret in the adversarial regime and $\mathcal{O}(\sum_{i,\Delta_i>0}\frac{\log(n)}{\Delta_i})$ regret in the stochastic regime simultaneously. This is the first FTPL algorithm to approach the BOBW guarantee in the semi-bandit setting when \( m \leq d/2 \). Technically, first, we use the standard analysis framework for FTRL algorithms (originally introduced by \cite{lattimore2020bandit}), and extend it to cases where the convex hull of the action set lacks interior points—i.e., to \( m \)-set semi-bandits—thereby simplifying \cite{pmlr-v201-honda23a}’s proof. Second, we generalize \cite{pmlr-v201-honda23a}’s analytical techniques and handle the challenges posed by the complex structure of arm-selection probabilities in \( m \)-set semi-bandits. Moreover, by establishing lower bounds, we demonstrate that our current approach has been pushed to its limit—namely, the $\log(d)$ and $\frac{m^{5/6}}{d^{1/2}}$ factors cannot be removed. Any further improvement would likely require adopting a different and more challenging line of analysis.

\subsection{Related Works}
\paragraph{FTPL}
The FTPL algorithm was originally introduced by \cite{a8076034-8fcb-3a02-aa25-0eb6a52176a4} in game theory and later rediscovered and formalized by \cite{KALAI2005291}. FTPL has since gained significant attention for its computational efficiency and adaptability across various online learning scenarios, including MAB \citep{abernethy2015fightingbanditsnewkind}, linear bandits \citep{10.1007/978-3-540-27819-1_8}, MDP bandits \citep{dai2022followtheperturbedleaderadversarialmarkovdecision}, combinatorial semi-bandits \citep{pmlr-v40-Neu15,neu2016importanceweightingimportanceweights} and Differential Privacy \citep{JMLR:v25:21-1267}. However, in MAB, due to the
complicated expression of the arm-selection probability in FTPL, it remains a open problem \citep{kim2019optimalityperturbationsstochasticadversarial} for a long time that dose there exist a perturbation achieve the optimal regret bound $\mathcal{O}(\sqrt{nd})$ in the adversarial setting, which had been already achieved by FTRL policies \citep{audibert:hal-00834882}. \cite{kim2019optimalityperturbationsstochasticadversarial} conjectured that the corresponding perturbations
should be of Fr\'echet-type tail distribution and it was shown to be true by \cite{pmlr-v201-honda23a, pmlr-v247-lee24a}.

\paragraph{BOBW}
Following the influential work of \cite{pmlr-v23-bubeck12b}, a broad line of research has explored BOBW algorithms across diverse online learning settings. These include, but are not limited to, MAB \citep{pmlr-v89-zimmert19a}, the problem of prediction with expert advice \citep{derooij2013followleadercanhedge, pmlr-v35-gaillard14,luo2015achievingparametersadaptivenormalhedge}, linear bandits \citep{ito2023bestofthreeworldslinearbanditalgorithm,kong2023bestofthreeworldsanalysislinearbandits}, 
dueling bandits \citep{pmlr-v162-saha22a}, contextual bandits \citep{pmlr-v238-kuroki24a}, episodic Markov decision processes \citep{jin2021bestworldsstochasticadversarial} and especially, combinatorial semi-bandits \citep{wei2018adaptivealgorithmsadversarialbandits,zimmert2019beating,NEURIPS2021_15a50c8b,pmlr-v206-tsuchiya23a}.

\section{Preliminaries}
In this section, we formulate the problem and introduce the FTPL policy.

\subsection{The Problem Setting and Notation}
We consider the $m$-set combinatorial semi-bandit problem with action set  
$
\mathcal{A} = \{ a \in \{0,1\}^d : \|a\|_1 = m \},
$  
where each action selects a subset of $m$ arms and $d\ge 2$. At each round \( t = 1, 2, \ldots \), the learner chooses an action \( A_t \in \mathcal{A} \), while the environment generates a loss vector \( \ell_t \in [0,1]^d \). The learner incurs loss \( \langle A_t, \ell_t \rangle \) and observes semi-bandit feedback \( o_t = A_t \odot \ell_t \), i.e., the losses for the chosen arms only. The goal is to minimize the pseudo-regret 
$
\operatorname{Reg}_n := \mathbb{E}\left[\sum_{t=1}^n \langle A_t - a_\star, \ell_t \rangle\right],
$  
where \( a_\star \in \arg\min_{a \in \mathcal{A}} \mathbb{E}\left[\sum_{t=1}^n \langle a, \ell_t \rangle\right] \) is the optimal fixed action in hindsight. In the adversarial setting, the loss vectors \( \ell_t \) may be arbitrary and adaptive. In the stochastic setting, they are i.i.d. samples from a fixed but unknown distribution \( \mathcal{D} \). Let \( \nu_i := \mathbb{E}_{\ell \sim \mathcal{D}}[\ell_i] \) denote the expected loss of arm \( i \). Define the suboptimality gap of arm \( i \) as  $
\Delta_i := \left(\nu_i - \max_{\mathcal{I} \subset \{1, \dots, d\}, |\mathcal{I}| < m} \min_{j \notin \mathcal{I}} \nu_j\right)^+
$  (A less formal way to put it is: the gap from the $m$-th smallest value.),
and the minimum gap as \( \Delta := \min_{i : \Delta_i > 0} \Delta_i \). Here  $(z)^+ = z \vee 0  :=\max(z, 0)$.

To analyze regret, for any $\lambda =(\lambda_1, \ldots, \lambda_d)^T \in\R^d$, we let $\lambda_i$ be the $\sigma_i(\lambda)$-th smallest among the $\lambda_i$ (ties are broken arbitrarily) and $\underline{\lambda}_i:=
 (
\lambda_i-\max\limits_{\mathcal{I},|\mathcal{I}|<m}\min\limits_{j\notin \mathcal{I}}\lambda_j
 )^+
$. 
% We also denote that $\widetilde{\sigma}_i(\lambda):=(\sigma_i(\lambda)-m)^++1.$
We denote by $\mathscr{F}_t$  the filtration $\sigma(A_1,o_1,K_1, \ldots, A_t,o_t)$, and by $\mathbf{1}$ the all-one vector. Let $a \wedge b =\min(a, b)$ and $a \vee b =\max(a, b)$.

We consider the Fr\'echet distribution  with shape parameter 2 (denoted \( \mathcal{F}_2 \)), the density and CDF of which are  
\[
f(x) = 2x^{-3}e^{-1/x^2}, \quad F(x) = e^{-1/x^2}, \quad x \geq 0,
\]  
respectively. 
In the following, ``Fr\'echet'' refers to this distribution without specifying the parameter.

\subsection{FTPL Policy}
We study the Follow-The-Perturbed-Leader (FTPL) algorithm (Algorithm \ref{alg: FTPL}), which selects actions based on a perturbed cumulative estimated loss  
$
\hat{L}_t = \sum_{s=1}^{t-1} \hat{\ell}_s.
$  
At round \( t \), the learner pulls the $m$ arms that rank among the $m$ smallest estimated loss with random perturbation $r_t/\eta_t$, where \( r_t \in \mathbb{R}^d \) has i.i.d. components drawn from the Fr\'echet distribution \( \mathcal{F}_2 \), and \( \eta_t = \mathcal{O}(t^{-1/2}) \) is the learning rate.  The probability of selecting arm \( i \) given \( \hat{L}_t \) is 
$
w_{t,i} = \phi_i(\eta_t \hat{L}_t),
$  
where for \( \lambda \in \R^d \),
\begin{equation}\label{defi phi}
    \phi_i(\lambda)=\P\{
    r_i-\lambda_i \mbox{ is among the top } m \mbox{ largest values in } r_1-\lambda_1, \ldots, r_d-\lambda_d
    \}.
\end{equation}
Then by Lemma \ref{phii=2v}, we have 
$\phi_i(\lambda)=2V_{i,3}(\lambda)$, where
\[ V_{i,N}(\lambda):=\int_{-\lambda_i}^{\infty} \frac{e^{-1/(x{+}\lambda_i)^2}}{(x{+}\lambda_i)^N}  \sum_{s=0}^{m-1} \sum_{\mathcal{I} \subseteq  \{1,\dots,d\} \setminus \{i\}, |\mathcal{\mathcal{I}}|=s} \bigg[ \prod_{q \in \mathcal{I}} (1 {-} F(x{+}\lambda_q)) \prod_{q \notin \mathcal{I}, q \neq i} F(x{+}\lambda_q) \bigg] \d x.\] 
We denote the true cumulative loss as \( L_t = \sum_{s=1}^{t-1} \ell_s \). For convenience, we also denote the vector function $\phi$ as $(\phi_1,\cdots,\phi_d)$ and $w_t$ as $(w_{t,1},\cdots,w_{t,d}).$

\begin{algorithm}[t]
   \caption{FTPL wit geometric resampling for $m$-set Semi-bandits}
   \label{alg: FTPL}
   \DontPrintSemicolon
   \SetAlgoLined
   \SetKwInOut{Initialization}{Initialization}
   \Initialization{$\hat{L}_1 = 0$}
   \For{$t=1, \ldots, n$}{
      Sample $r_t = (r_{t,1}, \ldots, r_{t,d})$ i.i.d. from $\mathcal{F}_2$.
      
      Play $A_t=\operatorname{argmin}\limits_{a \in\mathcal{A}}\agp{\hat{L}_{t}-r_{t} / \eta_t,a}$.
      
      Observe $o_t=A_t \odot \ell_t$. \\
      \For{$i=1,\ldots,d$}{
      %\If{$A_{t,i}=1$}{
      Set $K_{t,i}:=0$.
      
      \Repeat{$A'_{t,i}=1$}{
         $K_{t,i}:= K_{t,i}+1$.  \tcp*{geometric resampling}
         
         Sample $r' = (r_1', \ldots, r_d')$ i.i.d. from $\mathcal{F}_2$.

         $A'_t = \operatorname{argmin}\limits_{a \in\mathcal{A}}\agp{\hat{L}_{t}-r' / \eta_t,a}$.
      }
      %}
      Set $\widehat{w_{t,i}^{-1}} := K_{t,i}$, $\hat{\ell}_{t,i}=o_{t,i} \widehat{w_{t,i}^{-1}}$, and $\hat{L}_{t+1,i} := \hat{L}_{t,i} + \hat{\ell}_{t,i}$.
      }
   }
\end{algorithm}

\paragraph{Geometric Resampling}
In FTRL policies, Importance Weighted (IW) estimators are commonly used, where  
$
\hat{\ell}_{t,i} = \frac{\ell_{t,i} A_{t,i}}{w_{t,i}}, \text{ for } i = 1, \dots, d.
$
However, in FTPL algorithms, the action probabilities \( w_{t,i} \) are often hard to compute directly. To address this, the geometric resampling technique \citep{neu2016importanceweightingimportanceweights} is frequently employed. This method replaces \( w_{t,i}^{-1} \) with an unbiased estimator \( \widehat{w_{t,i}^{-1}} \). Specifically, after selecting action \( A_t \) and observing outcome \( o_t \), for each \( i = 1, \dots, d \), we repeatedly resample \( r' = (r_1', \ldots, r_d') \) i.i.d. from $\mathcal{F}_2$ and compute $A'_t = \operatorname{argmin}\limits_{a \in\mathcal{A}}\agp{\hat{L}_{t}-r' / \eta_t,a}$ until $A'_{t,i}=1$, i.e., arm \( i \) is ``selected''. Let \( K_{t,i} \) be the number of such resamples; then by the properties of the geometric distribution,  
$
\mathbb{E}[K_{t,i}] = \frac{1}{w_{t,i}},
$  
so we define \( \widehat{w_{t,i}^{-1}} := K_{t,i} \). To reduce computation, we only need to compute \( K_{t,i} \) for arms actually selected by \( A_t \) \citep{pmlr-v201-honda23a}. Since \( A_{t,i} = 0 \) implies \( \hat{\ell}_{t,i} = 0 \), the remaining estimates can be omitted.

\paragraph{Viewing as Mirror Descent}
FTPL can be interpreted as Mirror Descent \citep{abernethy2015fightingbanditsnewkind,lattimore2020bandit}. For all $\lambda\in\R^d$, let
\begin{equation}\label{definitionofphi}
    \begin{aligned}
    \Phi(\lambda)=&\E[\max\limits_{a\in\mathcal{A}}\agp{r+\lambda,a}
    ]\\
    =&\sum_{i=1}^d\E[(r_i+\lambda_i)\cdot\1_{\{\text{
    $r_i+\lambda_i$ is among the top $m$ largest values in $r_1+\lambda_1, \ldots,r_d+\lambda_d$
    }\}}].
\end{aligned}
\end{equation}
Then, by exchanging expectation and the derivation  (or see Lemma \ref{potentialphi}), it is clear that $\nabla\Phi(\lambda)=\phi(-\lambda)$ and $\Phi(\lambda)$ is convex. Consider the Fenchel dual of $\Phi$, $
        \Phi^*(u)=\sup_{x\in\R^d} \agp{x,u}-\Phi(x).
        $
    Then FTPL can be regarded as Mirror Descent with potential $\Phi^*$, because $w_t=\phi(\eta_t\hat{L}_{t})=\nabla\Phi(-\eta_t\hat{L}_{t}).$ However, it is worth noting that \(\nabla\Phi^*(w_t) = -\eta_t \hat{L}_t\) generally does not hold in this case, because for all $t\in\R, \phi(\lambda+t\mathbf{1})=\phi(\lambda)$ by its definition, and then \(\nabla\Phi\) is obviously not invertible.

\section{Main Results}
In this section, we present our main theoretical results, including the regret bounds and our new analyses of the regret decomposition.

%\subsection{Regret Bounds}

\begin{Thm}\label{adv}
    In the adversarial setting, Algorithm \ref{alg: FTPL} with learning rate $\eta_t=1/\sqrt{t}$ satisfies
    \[
    \operatorname{Reg}_n=\mathcal{O}\left(\sqrt{nm}(\sqrt{d\log(d)}+m^{5/6})\right).
    \]
\end{Thm}
\noindent
The proof is given in Section \ref{subadv}. Furthermore, Appendix \ref{lowertight} provides lower bounds showing that our current method (Section \ref{stbt}) is essentially tight—indicating that the $\log(d)$ and $\frac{m^{5/6}}{d^{1/2}}$ factors are inherent to our analysis. Thus, removing them would likely require fundamentally different and more sophisticated techniques.

In the stochastic setting, we assume that there are at most $m$ arms with $\Delta_i=0$. In other words, we assume the uniqueness of the optimal action $a_\star$. This is a common assumption in BOBW problems \citep{pmlr-v89-zimmert19a,zimmert2019beating, pmlr-v201-honda23a}. 
\begin{Thm}\label{sto}
    In the stochastic setting, if the optimal action is unique, then Algorithm \ref{alg: FTPL} with learning rate $\eta_t=1/\sqrt{t}$ satisfies
    \[
    \operatorname{Reg}_n=\mathcal{O}\left(\sum_{i,\Delta_i>0}\frac{\log(n)}{\Delta_i}\right)+\mathcal{O}\left(\frac{1}{\Delta}(m^2d\log(d)+m^{11/3}+md^2)\right),
    \]
    where $\Delta:=\min_{i,\Delta_i>0}\Delta_i$.
\end{Thm}
\noindent
Its proof can be found in Appendix \ref{pro sec}. Therefore, FTPL with Fr\'echet perturbations approaches BOBW when $m\leq d/2$. In addition, similar to \cite{pmlr-v89-zimmert19a,zimmert2019beating}, our algorithm adopts a simple time-decaying learning rate schedule \(\eta_t = 1/\sqrt{t}\). Our results can be readily extended to a more general setting with \(\eta_t = c/\sqrt{t}\) for any \(c > 0\).
\subsection{Regret Decomposition}
We follow the standard FTRL analysis framework for FTPL, originally by \citet[Theorem 30.4]{lattimore2020bandit}, extending it to $m$-set semi-bandits where the convex hull of the action set $\mathcal{A}$ has no interior points and hence \(\nabla\Phi\) and \(\nabla\Phi^*\) are not inverses of each other. For convenience, in the following, let $\eta_0=+\infty$. Then the regret can be decomposed in the following way:
\begin{Lem}\label{regret deco}
    \begin{equation*}
        \begin{aligned}
\operatorname{Reg_n}\leq&\underbrace{\E\left[\sum_{t=1}^n 
    \agp{\hat{\ell}_t,\phi(\eta_t \hat{L}_t)-\phi(\eta_t \hat{L}_{t+1})}\right]}_{\text{\rm Stability Term}}
    +\underbrace{\sum_{t=1}^n \left(\frac{1}{\eta_{t}}-\frac{1}{\eta_{t-1}}\right)\E\left[
    \Phi^*(a_\star)-\Phi^*(w_t)
    \right]}_{\text{\rm Penalty Term}}.
    \end{aligned}
    \end{equation*}
\end{Lem}
\noindent
Its proof is deferred in Appendix \ref{proof regret decom}. For the penalty term, we need the following result, whose proof can be found in Appendix \ref{proof phistar}.
\begin{Lem}\label{phistar}
    For all $\lambda\in\R^d$, let $a=\nabla\Phi(\lambda)$. Then $\Phi^*(a)=-\E[\agp{r,A}]$, where $A=\argmax_{a\in\mathcal{A}}\agp{r+\lambda,a}$.
    %and $\E[A]=a$
 Furthermore, for all $a\in\mathcal{A}$, we have $\Phi^*(a)\leq -\E[\agp{r,a}].$
\end{Lem}
\noindent
Combining Lemmas \ref{regret deco} and \ref{phistar}, we have
\begin{equation*}
        \begin{aligned}
\operatorname{Reg_n}\leq&\underbrace{\E\left[\sum_{t=1}^n 
    \agp{\hat{\ell}_t,\phi(\eta_t \hat{L}_t)-\phi(\eta_t \hat{L}_{t+1})}\right]}_{\text{\rm Stability Term}}
    +\underbrace{\sum_{t=1}^n \left(\frac{1}{\eta_{t}}-\frac{1}{\eta_{t-1}}\right)\E\left[
    \agp{r_t,A_t-a_\star}
    \right]}_{\text{\rm Penalty Term}},
    \end{aligned}
    \end{equation*}
    which is a stronger result and simplifies the proof compared to those of \cite{pmlr-v201-honda23a, pmlr-v247-lee24a}, where they still need to bound $\agp{\hat{\ell}_t,\phi(\eta_t \hat{L}_t)-\phi(\eta_{t+1} \hat{L}_{t+1})}$ for the stability term.
    \begin{Rema}
        Furthermore, by Generalized Pythagoras Identity (Lemma \ref{xyz}), for the stability term, one can obtain a tighter upper bound $\sum_{t=1}^n 
    \frac{1}{\eta_t}\E[D_\Phi(-\eta_t\hat{L}_{t+1},-\eta_t \hat{L}_t)]$, which is more popular in the analyses of FTRL policies and usually approximated by the sum of $\eta_t\E[\|\hat{\ell}_t\|^2_{\nabla^2\Phi(-\eta_t\hat{L}_t)}]$. However, such an approximate relationship is difficult to establish in FTPL because  $\nabla^2\Phi(-\eta_t\hat{L}_{t+1})$ and $\nabla^2\Phi(-\eta_t\hat{L}_{t})$ may not be close enough.
    \end{Rema}

\section{Proof Outline}
This section begins with analyses for the stability term and the penalty term, followed by a proof for Theorem \ref{adv} and a sketch for Theorem \ref{sto}, whose details can be found in Appendix \ref{pro sec}.  
Although our analysis follows the framework in \cite{pmlr-v201-honda23a}, directly applying their approach fails in the \(m\)-set semi-bandit setting due to the intricate structure of the arm selection probabilities.

\subsection{Stability Term}\label{stbt}
For the stability terms, informally, we will show that
\[
\E_{t-1}[\agp{\hat{\ell}_t,\phi(\eta_t \hat{L}_t)-\phi(\eta_t \hat{L}_{t+1})}]\lesssim
% -\eta_t\sum_{i=1}^d\hat{\ell}_{t,i}^2\frac{\partial}{\partial \lambda_i} \phi_i(\eta_t \hat{L}_t)=\eta_t\sum_{i=1}^d\ell_{t,i}^2 A_{t,i}\frac{-\frac{\partial}{\partial \lambda_i} \phi_i(\eta_t \hat{L}_t)}{\phi_i^2(\eta_t \hat{L}_t)}=
\eta_t\sum_{i=1}^d\frac{-\frac{\partial}{\partial \lambda_i} \phi_i(\eta_t \hat{L}_t)}{\phi_i(\eta_t \hat{L}_t)},
\]
and hence, the key component of the analysis lies in bounding the quantity  
$
- \frac{\frac{\partial}{\partial \lambda_i} \phi_i(\lambda)}{\phi_i(\lambda)},
$  
which, by the definition, is upper bounded by \(\frac{3V_{i,4}(\lambda)}{V_{i,3}(\lambda)}\). However, each \(V_{i,N}(\lambda)\) is a sum over many terms. To effectively bound this ratio, our strategy is to apply a union bound over all individual terms \(\frac{V_{i,4}^{\mathcal{I}}(\lambda)}{V_{i,3}^{\mathcal{I}}(\lambda)}\) such that $|\mathcal{I}|<m$ and $i\notin\mathcal{I}$, where we define  
\begin{equation}\label{vini}
    V_{i,N}^{\mathcal{I}}(\lambda) := \int_{-\lambda_i}^{\infty} \frac{1}{(x+\lambda_i)^N} e^{-1/(x+\lambda_i)^2} \prod_{q \in \mathcal{I}} (1 - F(x+\lambda_q)) \prod_{q \notin \mathcal{I},\ q \neq i} F(x+\lambda_q)\, \mathrm{d}x.
\end{equation}
To this end, we require the following generalization of \cite{pmlr-v201-honda23a}'s result, whose proof can be found in Appendix \ref{proof of mono2}.
\begin{Lem}\label{monontone2}
    For any $\mathcal{I}\subseteq \{1,\cdots,d\}$, $i\notin\mathcal{I}$, 
    %which doesn't contain $i$, 
    $\lambda\in\R^{d}$ such that $\lambda_i\ge 0$
    % such that for all $j\notin\mathcal{I}$, $\lambda_j\ge 0$
    and any $N\ge 3$, let
    \[
    J_{i,N,\mathcal{I}}(\lambda):=\int_{0}^{\infty} \frac{1}{(x+\lambda_i)^N}  \prod_{q \in \mathcal{I}} (1 - F(x+\lambda_q)) \prod_{q \notin \mathcal{I}} F(x+\lambda_q)\d x.
    \]
%     where $h(x)$ is an arbitrary non-negative 
% nice function (to exchange integral and the derivative)
 Then for all $k>0$, $\frac{J_{i,N+k,\mathcal{I}}(\lambda)}{J_{i,N,\mathcal{I}}(\lambda)}$ is increasing on $\lambda_q \ge 0$ for $q\notin\mathcal{I}$.
 % and $q\ne i$, while decreasing on $\lambda_q\ge 0$ for $q\in\mathcal{I}$.
\end{Lem}
\noindent
In the MAB setting (i.e., the case \( m = 1 \)), there are no \( 1 - F \) terms, and we can leverage monotonicity to let certain \( \lambda_q \) tend to infinity, making the corresponding \( F \) terms approach 1, which greatly simplifies the form of the ratio. However, for general \( m \), removing \( 1 - F \) terms would require sending the corresponding \( \lambda_q \) to negative infinity. This is not feasible, as the monotonicity does not hold generally. 
% only hold when \( \lambda_q \) is non-negative. 
To address this difficulty, we rely on the following result:
\begin{Lem}\label{hnk}
    For all $\mu\ge0$, $K\ge 1$, $ N\ge 3$ and $M\ge 1$, let $\mu_i\in\R$ for all $1\leq i\leq M$, and define
    \[
    H_{N}=\int_0^{+\infty} (x+\mu)^{-N}e^{-\frac{K}{(x+\mu)^2}}\prod_{i=1}^M\left(1-F(x+\mu_i)\right) \d x.
    \]
    For all $k\in\mathcal{N}^+$, we have
    \[
    \frac{H_{N+k}}{H_N}\leq C_{N,k}\left(\left(\frac{M}{K}\right)^{k/3}\wedge \mu^{-k}\right),
    \]
    where $C_{N,k}$ is a positive constant only depending on $N$ and $k$. Furthermore, if $K\ge M$, then we have
    \[
    \frac{H_{4}}{H_3}\leq C\left(\left(\frac{M}{K}\log\left(\frac{K}{M}+1\right)\right)^{1/2}\wedge \mu^{-1}\right),
    \]
    where $C$ is a positive constant.
\end{Lem}
\noindent
The proof of this lemma is tedious; it constitutes the most difficult part of the stability term analysis. Therefore, we defer it to Appendix \ref{proofhnk}. We also have lower bounds in Appendix \ref{lowertight}, showing that the logarithmic term and $M^{k/3}$ are inevitable. Based on Lemma \ref{hnk}, we have the following result, whose proof is deferred in Appendix \ref{proofstab1}.
\begin{Lem}\label{stab1}
    There exists $C>0$ such that for all $t\ge 1$ and $1\leq i\leq d$,
    \[
    \E\left[\hat{\ell}_{t,i}\left(\phi_i(\eta_t \hat{L}_t)-\phi_i(\eta_t \hat{L}_{t+1})\right)\mid\mathscr{F}_{t-1}\right]\leq
    C\cdot\underline{\hat{L}}_{t,i}^{-1}\wedge \eta_t\begin{cases}
        \sqrt{\frac{m\log(d)}{\sigma_i(\hat{L}_t)-m}} & \sigma_i(\hat{L}_t)>2m
        \\
        m^{1/3}%\sqrt{\log(m)+1} 
        & 
        \sigma_i(\hat{L}_t)\leq 2m.\end{cases}
    \]
\end{Lem}
As a direct corollary, we have:
\begin{Lem}\label{stab2}
    There exists $C>0$ such that for all $t\ge 1$,
    \[
    \E\left[\agp{\hat{\ell}_t,\phi(\eta_t \hat{L}_t)-\phi(\eta_t \hat{L}_{t+1})}\mid\mathscr{F}_{t-1}\right]\leq C\eta_t(\sqrt{md\log(d)}+m^{4/3}).
    \]
    \begin{proof}
        By Lemma \ref{stab1}, the left-hand side is less than
        \[
        C\eta_t\left(2m^{4/3}+\sum_{i,\sigma_i(\hat{L}_t)>2m}\sqrt{\frac{m\log(d)}{\sigma_i(\hat{L}_t)-m}}\right)\leq C'\eta_t(\sqrt{md\log(d)}+m^{4/3}).
        \]
        %where we used that we used $m\leq d$ and Lemma \ref{sumof}.
    \end{proof}
\end{Lem}
\noindent
% While we do not establish a formal lower bound, our numerical experiments indicate that \(\max\limits_{1 \leq i \leq d,\lambda\in\R^d}\frac{V_{4,i}(\lambda)}{V_{3,i}(\lambda)}\) scales as \(\Omega\left(\sqrt{\frac{\log(d)}{d}}\right)\), suggesting that Fréchet perturbations may fall short of achieving the optimal regret bound.
Finally, we also need a different upper bound making use of $\hat{L}_t$ in the stochastic environment and the proof can be found in Appendix \ref{proofstab3}, which used a new technique compared to \cite{pmlr-v201-honda23a}. Their proof relies on the uniqueness of the optimal arm, while there are $m$ in the $m$-set semi-bandits.
\begin{Lem}\label{stab3}
    If 
    $\sum\limits_{i=m+1}^d (\eta_t\underline{\hat{L}}_{t,i})^{-2}<\frac{1}{2m}$, then
    \[
    \E\left[\agp{\hat{\ell}_t,\phi(\eta_t \hat{L}_t)-\phi(\eta_t \hat{L}_{t+1})}\mid\mathscr{F}_{t-1}\right]\leq C \sum\limits_{i=m+1}^d \left(\underline{\hat{L}}_{t,i}^{-1}+\eta_td w_{t,i}
    \right)+m 2^{-\frac{1}{2\eta_td}},
    \]
    where $C$ is an absolute positive constant.
\end{Lem}

\subsection{Penalty Term}
Then we present our analyses for the penalty term.
\begin{Lem}\label{pen1}
    For all $\lambda\in\R^{d}$, we have
    \[
    \Phi^*(a)-\Phi^*(\phi(\lambda))\leq 5\sqrt{md}.
    \]
    Furthermore, if $a=\argmin_{a'\in\mathcal{A}}\agp{a',\lambda}$,
    then
    \[
    \Phi^*(a)-\Phi^*(\phi(\lambda))\leq 2\sum_{1\leq i\leq d,\sigma_i(\lambda)>m}\underline{\lambda}_i^{-1}.
    \]
\end{Lem}
\noindent
The proof is given in Appendix \ref{proof pen1}. It is worth noting that the first part of the result stems from a key observation: if one draws \( d \) i.i.d. samples from the Fr\'echet distribution, then the expected sum of the top \( m \) largest values among them can be upper bounded by \( \mathcal{O}(\sqrt{md}) \). Then clearly, we have:
\begin{Lem}\label{pen2}
    For all $t\ge1$, we have
        \[
        \E\left[
    \Phi^*(a_\star)-\Phi^*(w_{t})
    \right]\leq 5\sqrt{md}.
        \]
    Furthermore, if 
    $\max\limits_{1\leq i\leq m}\hat{L}_{t,i}\leq\min\limits_{m+1\leq i\leq d}\hat{L}_{t,i}$
    and $a_\star=(\underbrace{1,\cdots,1}_{m \text{ of } 1},\underbrace{0,\cdots,0}_{d-m \text{ of } 0})$, then
    \[
        \E\left[
    \Phi^*(a_\star)-\Phi^*(w_{t})
    \mid\mathscr{F}_{t-1}\right]\leq 2\eta_t^{-1}\sum_{i=m+1}^d \underline{\hat{L}}_{t,i}^{-1}.
    \]
    % \begin{proof}
    %     It suffices to apply Lemma \ref{pen1}.
    % \end{proof}
\end{Lem}

\subsection{Proof for Theorem \ref{adv}}\label{subadv}
Combining Lemmas \ref{regret deco}, \ref{stab2} and \ref{pen2} with $\eta_t=1/\sqrt{t}$, we have
\[
\operatorname{Reg}_n\leq C\sum_{t=1}^n \eta_t(\sqrt{md\log(d)}+m^{4/3})+5\sum_{t=1}^n \left(\frac{1}{\eta_{t}}-\frac{1}{\eta_{t-1}}\right)\sqrt{md}\leq C'\sqrt{nm}(\sqrt{d\log(d)}+m^{5/6}),
\]
where we applied Lemma \ref{sumof}.

\subsection{Proof Sketch for Theorem \ref{sto}}
W.L.O.G., we assume that $\nu_1\leq\nu_2\leq\cdots\leq\nu_d$ and then $a_\star=(\underbrace{1,\cdots,1}_{m \text{ of } 1},\underbrace{0,\cdots,0}_{d-m \text{ of } 0})$. We apply the technique by \cite{pmlr-v201-honda23a} and hence define the event $A_t=\{\sum\limits_{i=m+1}^d (\eta_t\underline{\hat{L}}_{t,i})^{-2}<\frac{1}{2m}\}$. On one hand, by Lemma \ref{stab2}, \ref{stab3} and \ref{pen2}, one can show that
\[
\begin{aligned}
    \operatorname{Reg}_n\leq& \underbrace{\mathcal{O}\left(\sum_{t=1}^n\E\left[
\1_{\{A_t\}}\cdot \sum\limits_{i=m+1}^d \underline{\hat{L}}_{t,i}^{-1}
+\1_{\{A_t^c\}}\sqrt{\frac{m}{t}}(\sqrt{d\log(d)}+m^{5/6})
\right]\right)}_{\uppercase\expandafter{\romannumeral1}}\\
&+\underbrace{\mathcal{O}\left(\sum_{t=1}^n\E\left[
\sum_{i=m+1}^{d}\frac{d w_{t,i}}{\sqrt{t}}
\right]\right)}_{\uppercase\expandafter{\romannumeral2}}+\mathcal{O}\left(md^2\right).
\end{aligned}
\]
 On the other hand, using the fact that \[
\operatorname{Reg}_n\ge\sum_{t=1}^n\E\left[
\sum_{i=m+1}^d\Delta_i w_{t,i}
\right]\ge \underbrace{\Delta\sum_{t=1}^n\E\left[
\sum_{i=m+1}^d w_{t,i}
\right]}_{\uppercase\expandafter{\romannumeral4}},
\]
in Appendix \ref{pro sec} we will show that
\[
\operatorname{Reg}_n\ge \underbrace{\Omega\left(\sum_{t=1}^n\E\left[
\1_{\{A_t\}}\cdot t\sum\limits_{i=m+1}^d \Delta_i\underline{\hat{L}}_{t,i}^{-2}
+\1_{\{A_t^c\}}\cdot\frac{\Delta}{m}
\right]\right)}_{\uppercase\expandafter{\romannumeral3}}.
\]
Hence, with
$
\operatorname{Reg}_n=3\operatorname{Reg}_n-2\operatorname{Reg}_n\leq (3\uppercase\expandafter{\romannumeral1}-\uppercase\expandafter{\romannumeral3})+(3\uppercase\expandafter{\romannumeral2}-\uppercase\expandafter{\romannumeral4})+\mathcal{O}\left(md^2\right),
$ one can get the logarithmic result by noting that $\underline{\hat{L}}_{t,i}^{-1}-t\Delta_i\underline{\hat{L}}_{t,i}^{-2}=\mathcal{O}(\frac{1}{t\Delta_i})$ and $\sqrt{\frac{m}{t}}(\sqrt{d\log(d)}+m^{5/6})$ and $\frac{dw_{t,i}}{\sqrt{t}}$ are less than $\frac{\Delta}{m}$ and $\Delta w_{t,i}$ respectively when $t$ is large enough. Details can be found in Appendix \ref{pro sec}.
% which is less than
% \[
% \begin{aligned}
%     \mathcal{O}\left(\sum_{t=1}^n\E\left[
% \1_{\{A_t\}}\cdot \sum\limits_{i=m+1}^d \left(\underline{\hat{L}}_{t,i}^{-1}-t\Delta_i\underline{\hat{L}}_{t,i}^{-2}\right)
% +\1_{\{A_t^c\}}\sqrt{\frac{md}{t}}
% \right]\right)+\mathcal{O}\left(\sum_{t=1}^n\E\left[
% \sum_{i=m+1}^{d}\frac{ w_{t,i}}{\sqrt{t}}
% \right]\right)+\mathcal{O}\left(m\right)+\mathcal{O}(1).
% \end{aligned}
% \]
\begin{figure}[htbp]
    \centering
    \begin{minipage}[b]{0.9\textwidth} % [b] 表示底部对齐，0.8\textwidth 是 minipage 的宽度
        \centering
        \includegraphics[width=0.9\textwidth]{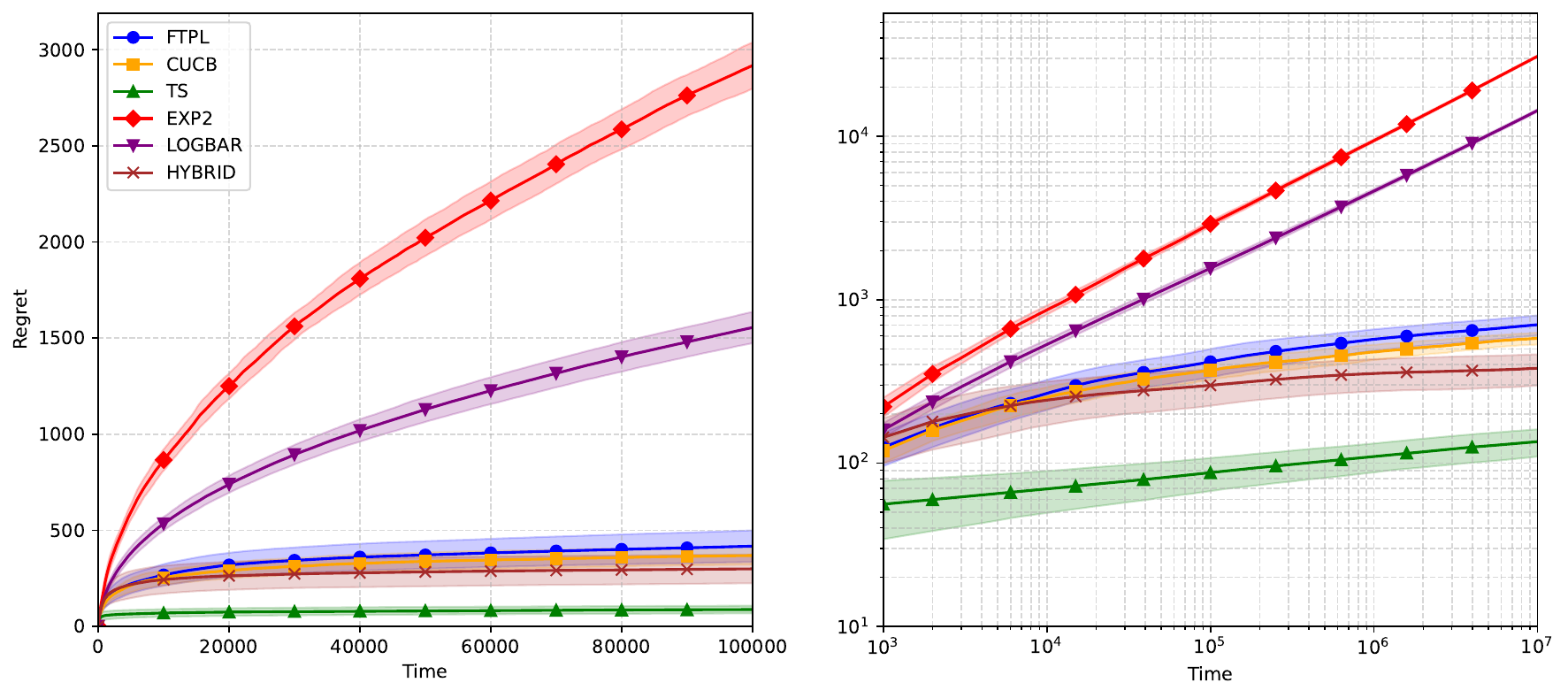}
        %\captionof{subfigure}{第一张子图的简短描述} % 使用 captionof 来添加子图的标题
        \label{fig:sub1-minipage}
    \end{minipage}
    \vfill % 在两个 minipage 之间添加垂直间距
    \begin{minipage}[b]{0.9\textwidth}
        \centering
        \includegraphics[width=0.9\textwidth]{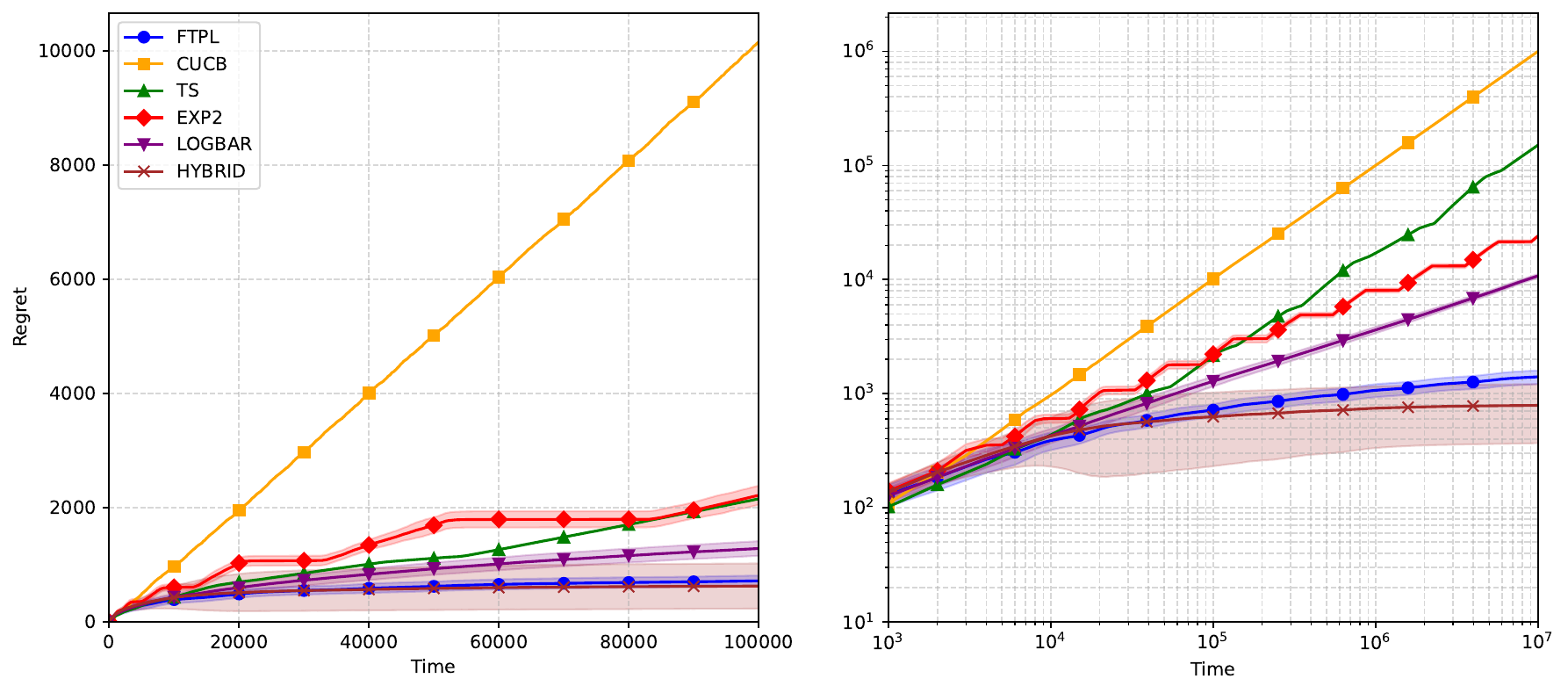}
        %\captionof{subfigure}{第二张子图的简短描述}
        \label{fig:sub2-minipage}
    \end{minipage}
    \caption{Comparisons of our algorithm FTPL and several
existing algorithms. The left side is in linear scale and the right is in log-log scale.}
    \label{results}
\end{figure}

\section{Experiments}
In this section, we evaluate the empirical performance of FTPL and several benchmark algorithms on the $m$-set semi-bandit problem. We compare our method against five established baselines: for the stochastic setting, we include COMBUCB \citep{kveton2015tightregretboundsstochastic} and THOMPSON SAMPLING \citep{pmlr-v32-gopalan14}; for the adversarial setting, we use EXP2 \citep{audibert2014regret} and LOGBARRIER \citep{luo2018efficientonlineportfoliologarithmic}, corresponding to FTRL with generalized Shannon entropy and log-barrier regularizers, respectively. We also compare against the BOBW algorithm—FTRL with a hybrid regularizer (hereafter referred to as “hybrid”) \citep{zimmert2019beating}—in both settings.

Following \citep{zimmert2019beating}, we run experiments on a specific instance of the $m$-set semi-bandit with parameters $d = 10$, $m = 5$, and $n = 10^7$. The loss for arm $i$ at time $t$ has mean $\nu_{ti}$, and the realized loss is $0$ with probability $1-\nu_{ti}$ and $1$ with probability $\nu_{ti}$, independently across arms and time. 
% Since the hybrid algorithm is designed for losses in $[-1, 1]$, we rescale the bandit feedback by multiplying it by 2 and subtracting 1 before inputting it into the algorithm. 
In the stochastic environment, the losses are generated from a stationary distribution where the mean loss for arm $i$ at time $t$ is given by $\nu_{ti} = \frac{1}{2} - \Delta$ if $i \leq 5$, and $\nu_{ti} = \frac{1}{2} + \Delta$ otherwise, with $\Delta = 0.1$. In the adversarial environment, we employ the adversarial setting detailed in \cite{zimmert2019beating}, a framework with numerous practical applications. This setting divides the time horizon into phases:
$
1, \dots, n_1, n_1 + 1, \dots, n_2, \dots, n_{k-1}, \dots, n.
$
The duration of phase $s$ is $N_s = 1.6^s$, and the mean losses are configured as follows:
\[
\nu_{ti} =
\begin{cases}
\frac12-\frac\Delta4 \pm (\frac12 - \frac\Delta4) & \text{if } i \leq 5, \\
\frac12+\frac\Delta4 \pm (\frac12 - \frac\Delta4) & \text{otherwise},
\end{cases}
\]
where $\Delta=0.1$ and in $\pm$, $+$ is used if time $t$ falls within an odd-numbered phase, and $-$ otherwise. We sample a sequence of $n$ loss vectors from the above setting and fix it as our adversarial environment, then run the algorithms to be compared on this fixed sequence. Across all experiments, we estimated the pseudo-regret using 20 repetitions. The resulting average pseudo-regret for each algorithm over time is presented in Figure \ref{results}. Our experiments are conducted on a server with 4 NVIDIA RTX 4090 GPUs and Intel(R) Xeon(R) Gold 6132 CPU @ 2.60GHz.

\section{Concluding Remarks}
To summarize, we have shown that FTPL with Fréchet perturbations achieves both \(\mathcal{O}(\sqrt{nm}(\sqrt{d\log(d)}+m^{5/6}))\) regret in the adversarial regime and \(\mathcal{O}(\sum_{i,\Delta_i>0}\frac{\log(n)}{\Delta_i})\) regret in the stochastic regime. This makes it the first FTPL algorithm to approach the Best-of-Both-Worlds (BOBW) guarantee in the $m$-set semi-bandit setting when \( m \leq d/2 \). Our analysis has been built upon the standard FTRL framework, which we extend to accommodate the lack of interior points in the convex hull of the \( m \)-set action space. In doing so, we simplify and partially extend the arguments of \cite{pmlr-v201-honda23a}, and attempt to address the technical challenges arising from the intricate structure of arm-selection probabilities in the semi-bandit setting.

An important open question is whether a sharper upper bound on $\frac{V_{4,i}}{V_{3,i}}$ can be established to eliminate the $\log(d)$ and $\frac{m^{5/6}}{d^{1/2}}$ factors in the regret bound, thereby enabling FTPL to achieve the BOBW guarantee. Appendix \ref{lowertight} suggests that Lemma \ref{hnk} is already tight, meaning that these factors cannot be removed. Therefore, obtaining a tighter bound on $\frac{V_{4,i}}{V_{3,i}}$ is not possible through bounding the term-wise ratio; instead, one must analyze the ratio of the full summations directly, which is substantially more challenging. Moreover, it has been shown that FTRL algorithms \citep{zimmert2019beating} can achieve the BOBW guarantee even in the regime where \( m \ge d/2 \). Whether there exists an FTPL algorithm capable of matching this performance remains an intriguing open problem. Promising future directions include extending our analysis to more general Fr\'echet distributions \citep{pmlr-v247-lee24a}, or investigating broader classes of 
the semi-bandit settings.

\section*{Acknowledgments}
We would like to thank a person, who wishes to remain anonymous, for pointing out an issue in the previous version of Lemma \ref{monontone2}: specifically, the second part of the lemma incorrectly claimed a monotonicity property with respect to $\lambda_q$ for $q \in \mathcal{I}$. This statement was incorrect and has been removed in the updated version. To address this gap, we have introduced Lemma \ref{hnk}, which does not require any assumptions on $\mu_i$, and thus eliminates the need for the previously claimed monotonicity.

With these modifications, our regret bound in the adversarial setting incurs additional $\log(d)$ and $\frac{m^{5/6}}{d^{1/2}}$ factors. In the stochastic setting, the logarithmic regret rate remains correct, but the additive term becomes slightly larger. Despite these changes, our results still demonstrate that FTPL approaches the BOBW guarantee in the $m$-set semi-bandit problems.

\bibliography{ref}
\bibliographystyle{plainnat}
\appendix
\newpage
\renewcommand{\appendixpagename}{\centering \LARGE Appendix}
\appendixpage

\startcontents[section]
\printcontents[section]{l}{1}{\setcounter{tocdepth}{2}}
\newpage
\section{Proof for Theorem \ref{sto}}\label{pro sec}
W.L.O.G., we assume that $\nu_1\leq\nu_2\leq\cdots\leq\nu_d$ and then $a_\star=(\underbrace{1,\cdots,1}_{m \text{ of } 1},\underbrace{0,\cdots,0}_{d-m \text{ of } 0})$. Define the event $A_t=\{\sum\limits_{i=m+1}^d (\eta_t\underline{\hat{L}}_{t,i})^{-2}<\frac{1}{2m}\}$ and $w_\star^t=\P\{A_t=a_\star\mid\mathscr{F}_{t-1}\}.$

Our plan is to apply the self-bounding constrain technique by \cite{pmlr-v201-honda23a,pmlr-v89-zimmert19a}. We first derive the upper bound. On the one hand, if $\hat{L}_t$ satisfies $A_t$, which implies that $\max\limits_{1\leq i\leq m}\hat{L}_{t,i}\leq\min\limits_{m+1\leq i\leq d}\hat{L}_{t,i}$, then combining Lemma \ref{regret deco}, \ref{stab3} and \ref{pen2}, the regret in round $t$ should be bounded by
\begin{equation}\label{upper at}
    C \sum\limits_{i=m+1}^d \underline{\hat{L}}_{t,i}^{-1}
    + C\sum_{i=m+1}^{d}\frac{d w_{t,i}}{\sqrt{t}}+m2^{-\sqrt{t}/2d},
\end{equation}
where we 
% denoted that $w_t=\phi(\eta_t\hat{L}_t)$ and 
used that $\eta_{t+1}^{-1}-\eta_{t}^{-1}=\sqrt{t+1}-\sqrt{t}\leq\frac{1}{2\sqrt{t}}=\eta_t/2$ for the penalty term.
\noindent
On the other hand, if $\hat{L}_t$ doesn't satisfy $A_t$, similarly, by Lemma \ref{regret deco}, \ref{stab2} and \ref{pen2}, the regret in round $t$ is less than
\begin{equation}\label{upper atc}
    C \sqrt{\frac{m}{t}}(\sqrt{d\log(d)}+m^{5/6}).
\end{equation}
Putting \eq{upper at} and \eq{upper atc} together, one can get
\[
\operatorname{Reg}_n\leq \underbrace{C\sum_{t=1}^n\E\left[
\1_{\{A_t\}}\cdot \sum\limits_{i=m+1}^d \underline{\hat{L}}_{t,i}^{-1}
+\1_{\{A_t^c\}}\sqrt{\frac{m}{t}}(\sqrt{d\log(d)}+m^{5/6})
\right]}_{\uppercase\expandafter{\romannumeral1}}+\underbrace{C\sum_{t=1}^n\E\left[
\sum_{i=m+1}^{d}\frac{d w_{t,i}}{\sqrt{t}}
\right]}_{\uppercase\expandafter{\romannumeral2}}+Cmd^2,
\]
 where we applied Lemma \ref{2etat} for the last term.

We then show the lower bound. Clearly, we have
\begin{equation}\label{lower1}
    \operatorname{Reg}_n\ge
    \sum_{t=1}^n\E\left[\sum_{i=m+1}^d \Delta_i w_{t,i}\right]
    \ge \sum_{t=1}^n\E\left[\1_{\{A_t\}}\cdot\sum_{i=m+1}^d \Delta_i w_{t,i}+\1_{\{A_t^c\}}\cdot\Delta(1-w^t_{\star})\right]
    %\ge \Delta\sum_{t=1}^n\E[1-w^t_{\star}]
    ,
\end{equation}
where we applied Lemma \ref{wstarwti} for the second term. Then by Lemma \ref{wstarbound} and Lemma \ref{phiilower}, we have
\[
\operatorname{Reg}_n\ge \underbrace{C'\sum_{t=1}^n\E\left[
\1_{\{A_t\}}\cdot t\sum\limits_{i=m+1}^d \Delta_i\underline{\hat{L}}_{t,i}^{-2}
+\1_{\{A_t^c\}}\cdot\frac{\Delta}{m}
\right]}_{\uppercase\expandafter{\romannumeral3}},
\]
where $C'$ is an absolute positive constant. Besides, similar to \eq{lower1}, we also have
\[
\operatorname{Reg}_n\ge \underbrace{\Delta\sum_{t=1}^n\E\left[
\sum_{i=m+1}^d w_{t,i}
\right]}_{\uppercase\expandafter{\romannumeral4}}.
\]
Hence,
\[
\operatorname{Reg}_n=3\operatorname{Reg}_n-2\operatorname{Reg}_n\leq (3\uppercase\expandafter{\romannumeral1}-\uppercase\expandafter{\romannumeral3})+(3\uppercase\expandafter{\romannumeral2}-\uppercase\expandafter{\romannumeral4})+Cmd^2.
\]
For $3\uppercase\expandafter{\romannumeral1}-\uppercase\expandafter{\romannumeral3},$ it equals
\[
\sum_{t=1}^n\E\left[
\1_{\{A_t\}}\cdot\left( 3C\sum\limits_{i=m+1}^d \underline{\hat{L}}_{t,i}^{-1}-C't\sum\limits_{i=m+1}^d \Delta_i\underline{\hat{L}}_{t,i}^{-2}
\right)\right]
+\sum_{t=1}^n\E\left[
\1_{\{A_t^c\}}\cdot\left(3C\sqrt{\frac{m}{t}}(\sqrt{d\log(d)}+m^{5/6})-C'\frac{\Delta}{m}\right)
\right].
\]
For the first term, since $ax-bx^2\leq a^2/4b$ for $b>0$, then there exists $C''>0$ such that
\[
3C\sum\limits_{i=m+1}^d \underline{\hat{L}}_{t,i}^{-1}-C't\sum\limits_{i=m+1}^d \Delta_i\underline{\hat{L}}_{t,i}^{-2}\leq 
%\frac{C''(d-m)}{t\Delta}.
\sum_{i=m+1}^d\frac{C''}{t\Delta_i}.
\]
Note that $3C\sqrt{\frac{m}{t}}(\sqrt{d\log(d)}+m^{5/6})\leq C'\frac{\Delta}{m}$ after $\sqrt{t}\ge\frac{3Cm^{\frac32}(\sqrt{d\log(d)}+m^{5/6})}{C'\Delta}$, then we have
\begin{equation}
    \begin{aligned}
        3\uppercase\expandafter{\romannumeral1}-\uppercase\expandafter{\romannumeral3}&\leq 
        \sum_{t=1}^n\sum_{i=m+1}^d\frac{C''}{t\Delta_i}+\sum_{t=1}^{\frac{9C^2m^3(\sqrt{d\log(d)}+m^{5/6})^2}{C'^2\Delta^2}}3C\sqrt{\frac{m}{t}}(\sqrt{d\log(d)}+m^{5/6})\\
        &\leq \sum_{i=m+1}^d\frac{C_1\log(n)}{\Delta_i}+\frac{C_2m^2(d\log(d)+m^{5/3})}{\Delta},
    \end{aligned}
\end{equation}
where $C_1$ and $C_2$ are absolute positive constants and we used $(x+y)^2\leq 2(x^2+y^2) $ for positive $x$ and $y$.

Then it suffices to bound $3\uppercase\expandafter{\romannumeral2}-\uppercase\expandafter{\romannumeral4}$, which equals
\[
\sum_{t=1}^n\E\left[\sum_{i=m+1}^d\left(\frac{3Cdw_{t,i}}{\sqrt{t}}-\Delta w_{t,i}\right)\right].
\]
Similarly, $\frac{3Cdw_{t,i}}{\sqrt{t}}\leq\Delta w_{t,i}$ after $\sqrt{t}\ge \frac{3Cd}{\Delta}$. Hence,
\[
\sum_{t=1}^n\E\left[\sum_{i=m+1}^d\left(\frac{3Cdw_{t,i}}{\sqrt{t}}-\Delta w_{t,i}\right)\right]\leq m\sum_{t=1}^{\frac{9C^2d^2}{\Delta^2}}\frac{3Cd}{\sqrt{t}}\leq \frac{C_3 md^2}{\Delta},
\]
where $C_3$ is an absolute positive constant and we used that $\sum_{i=1}^d w_{t,i}=m$ by Lemma \ref{sum=m}. We complete the proof by putting everything together.

\section{Decomposition}\label{deco sec}
In this section, we give the detailed proof for the regret decomposition.
\begin{Lem}\label{decom2}
    Let $\ell_1,\cdots,\ell_n\in\R^{d}$ and $a_t=\phi(\eta_t L_t)$, where $\left(\eta_t\right)_{t=0}^{n}$ is decreasing with 
    % $\eta_n=\eta_{n+1}$ and
    $\eta_0=+\infty$ and $L_t:=\sum\limits_{s=1}^{t-1}\ell_s.$ Then for all $a\in\mathcal{A}$,
    \[
    \sum_{t=1}^n \agp{a_t-a,\ell_t}\leq\sum_{t=1}^n 
    % \frac{1}{\eta_t}D_\Phi(-\eta_tL_{t+1},-\eta_t L_t)
    \agp{\ell_t,\phi(\eta_t L_t)-\phi(\eta_t L_{t+1})}
    +\sum_{t=1}^n \left(\frac{1}{\eta_{t}}-\frac{1}{\eta_{t-1}}\right)\left(
    \Phi^*(a)-\Phi^*(a_{t})
    \right).
    \]
    \begin{proof}
    For convenience, let $\eta_{n+1}=\eta_n$ and $a_{n+1}=\phi(\eta_n L_{n+1})$.
    Note that $a_t=\nabla\Phi(-\eta_t L_t)$, then by Lemma \ref{xxstar},$-\eta_t L_{t}\in\pd\Phi^*(a_t)$, which implies that $a_t\in\argmin_{x\in\R^d} \Phi_t^*(x),$ where $\Phi_t^*(x):=\frac{\Phi^*(x)}{\eta_t}+\agp{x, L_t}.$ We then have
    \[
    \begin{aligned}
        \sum_{t=1}^n \agp{a_t-a,\ell_t}=&\sum_{t=1}^n \agp{a_t-a_{t+1},\ell_t}+\sum_{t=1}^n \agp{a_{t+1},\ell_t}-\sum_{t=1}^n \agp{a,\ell_t}\\
        =&\sum_{t=1}^n \agp{a_t-a_{t+1},\ell_t}+\sum_{t=1}^n\left(
        \Phi_{t+1}^*(a_{t+1})-\frac{\Phi^*(a_{t+1})}{\eta_{t+1}}-\left[\Phi_{t}^*(a_{t+1})-\frac{\Phi^*(a_{t+1})}{\eta_{t}}
        \right]
        \right)\\
        &-\sum_{t=1}^n\left(
        \Phi_{t+1}^*(a)-\frac{\Phi^*(a)}{\eta_{t+1}}-\left[\Phi_{t}^*(a)-\frac{\Phi^*(a)}{\eta_{t}}
        \right]
        \right)\\
        =&\sum_{t=1}^n \agp{a_t-a_{t+1},\ell_t}+\sum_{t=1}^n\left(
    \Phi_t^*(a_t)-\Phi_t^*(a_{t+1})\right)+\sum_{t=1}^n \left(\frac{1}{\eta_{t}}-\frac{1}{\eta_{t-1}}\right)\left(
    \Phi^*(a)-\Phi^*(a_{t})
    \right)\\
    &+\Phi_{n+1}^*(a_{n+1})-\Phi_{n+1}^*(a).
    \end{aligned}
    \]
    Since for all $a\in\mathcal{A}$, $\Phi^*_{n+1}(a_{n+1})\leq \Phi^*_{n+1}(a)$, we have
    \[
    \sum_{t=1}^n \agp{a_t-a,\ell_t}\leq \sum_{t=1}^n\left(\agp{a_t-a_{t+1},\ell_t}+
    \Phi_t^*(a_t)-\Phi_t^*(a_{t+1})\right)+\sum_{t=1}^n \left(\frac{1}{\eta_{t}}-\frac{1}{\eta_{t-1}}\right)\left(
    \Phi^*(a)-\Phi^*(a_{t})
    \right).
    \]
    Then by the definition,
    \[
    \begin{aligned}
        \Phi_t^*(a_t)-\Phi_t^*(a_{t+1})=-\frac1{\eta_t}\left(
    \Phi^*(a_{t+1})-\Phi^*(a_t)-\agp{a_{t+1}-a_t,-\eta_t L_t}
    \right)=-\frac1{\eta_t}D_{\Phi}(-\eta_t L_t,-\eta_{t+1} L_{t+1}),
    \end{aligned}
    \]
    where we used Lemma \ref{uv xy} by noting that $\nabla\Phi(-\eta_t L_t)=a_t$ and $\nabla\Phi(-\eta_{t+1} L_{t+1})=a_{t+1}$. Finally, by Lemma \ref{xyz} (taking $x=-\eta_t L_{t+1}, y=-\eta_{t+1}L_{t+1}$ and $z=-\eta_t L_{t}$), we have
    \[
    \agp{a_t-a_{t+1},\ell_t}-\frac1{\eta_t}D_{\Phi}(-\eta_t L_t,-\eta_{t+1} L_{t+1})\leq
    %\frac{1}{\eta_t}D_\Phi(-\eta_tL_{t+1},-\eta_t L_t)
    \agp{\ell_t,\nabla\Phi(-\eta_t L_t)-\nabla\Phi(-\eta_t L_{t+1})},
    \]
    since $D_\Phi(x,y)+D_\Phi(z,x)\ge 0$. We complete the proof by putting them together.
    \end{proof}
    \begin{Rema}
        The overall proof framework is based on \citet[Exercise 28.12]{lattimore2020bandit}, with the latter part inspired by \citet[Lemma 3]{zimmert2019connectionsmirrordescentthompson}.
    \end{Rema}
\end{Lem}
\subsection{Proof for Lemma \ref{regret deco}}\label{proof regret decom}
\begin{proof}
        Noting that $\E[A_t\mid\mathscr{F}_{t-1}]=\phi(\eta_t\hat{L}_t)$, we have
\begin{equation*}\label{decom1}
    \operatorname{Reg}_n=\mathbb{E}\left[\sum_{t=1}^n\left\langle \phi(\eta_t\hat{L}_t)-a_\star, \ell_t\right\rangle\right]=\mathbb{E}\left[\sum_{t=1}^n\left\langle \phi(\eta_t\hat{L}_t)-a_\star, \hat{\ell}_t\right\rangle\right].
\end{equation*}
Then it suffices to apply Lemma \ref{decom2}.
    \end{proof}

\section{Important Facts}\label{fact sec}
In this section, we present some important facts to be used in our analyses.
\begin{Lem}\label{potentialphi}
    For all $\lambda\in\R^d$, we have $\nabla\Phi(\lambda)=\phi(-\lambda)$ and $\Phi(\lambda)$ is convex over $\R^d$.
    \begin{proof}
        By \eq{definitionofphi}, since for all $1\leq i\leq d$, $\E|r_i|<+\infty$, one can exchange expectation and the derivative, then we have
        \[
        \frac{\pd}{\pd\lambda_i}\Phi(\lambda)=\E\left[\1_{\{\text{
    $r_i+\lambda_i$ is among the top $m$ largest values in $r_1+\lambda_1,\cdots,r_d+\lambda_d$
    }\}}\right]=\phi_i(-\lambda),
        \]
    because
    \[
    \frac{\pd}{\pd\lambda_i}\1_{\{\text{
    $r_i+\lambda_i$ is among the top $m$ largest values in $r_1+\lambda_1,\cdots,r_d+\lambda_d$
    }\}}=0,\,a.s.
    \]
    This shows that $\nabla\Phi(\lambda)=\phi(-\lambda)$. For convexity, it suffices to note that taking maximum and expectation keeps convexity.
    \end{proof}
\end{Lem}
\begin{Lem}\label{sum=m}
    For all $\lambda\in\R^d$, we have $\sum_{i=1}^d\phi_i(\lambda)=m$.
    \begin{proof}
        By the definition, we have
        \[
        \sum_{i=1}^d\phi_i(\lambda)=\E\left[\sum_{i=1}^d
        \1_{\left\{
        \text{
    $r_i-\lambda_i$ is among the top $m$ largest values in $r_1-\lambda_1,\cdots,r_d-\lambda_d$
    }
        \right\}}
        \right]=m.
        \]
    \end{proof}
\end{Lem}
\begin{Lem}\label{phii=2v}
    $\phi_i(\lambda)=2V_{i,3}(\lambda)$
    \begin{proof}
        Because
        \[
        \begin{aligned}
            \phi_i(\lambda)=\E_{r_i}&[\P\{\text{there exist at most $m-1$ of $r_1-\lambda_1,\cdots,r_{i-1}-\lambda_{i-1},$}\\
            &\text{$r_{i+1}-\lambda_{i+1},\cdots,r_d-\lambda_d$ that are larger than $x$}\mid r_i-\lambda_i=x\}],   
        \end{aligned}
        \]
        then it suffices to note that the conditional probability inside is just
        \[
        \sum_{s=0}^{m-1} \sum_{\mathcal{I} \subseteq  \{1,\dots,d\} \setminus \{i\}, |\mathcal{\mathcal{I}}|=s} \left[ \prod_{q \in \mathcal{I}} (1 - F(x+\lambda_q)) \prod_{q \notin \mathcal{I}, q \neq i} F(x+\lambda_q) \right].
        \]
    \end{proof}
\end{Lem}

\subsection{Proof for Lemma \ref{monontone2}}\label{proof of mono2}
 \begin{proof}
    We follow the proof by \cite{pmlr-v201-honda23a}. Let $Q(x)=h(x)(x+\lambda_i)^{-N}\prod_{q \in \mathcal{I}} (1 - F(x+\lambda_q)) \prod_{q \notin \mathcal{I}} F(x+\lambda_q)$. If $q\notin\mathcal{I}$, then
    \[
    \frac{\pd}{\pd\lambda_q}J_{i,N,\mathcal{I}}(\lambda)=2\int_0^{+\infty} (x+\lambda_q)^{-3}Q(x)\d x:=2J_{i,N,\mathcal{I}}^q(\lambda).
    \]
    Hence,
    \[
    \frac{\pd}{\pd\lambda_q}\frac{J_{i,N+k,\mathcal{I}}(\lambda)}{J_{i,N,\mathcal{I}}(\lambda)}=2\cdot\frac{J_{i,N+k,\mathcal{I}}^q(\lambda)J_{i,N,\mathcal{I}}(\lambda)-J_{i,N+k,\mathcal{I}}(\lambda)J_{i,N,\mathcal{I}}^q(\lambda)}{J_{i,N,\mathcal{I}}(\lambda)^2}.
    \]
    Note that
    \[
    \begin{aligned}
        J_{i,N+k,\mathcal{I}}^q(\lambda)J_{i,N,\mathcal{I}}(\lambda)=&\int\int_{x,y\ge 0}(x+\lambda_q)^{-3}(x+\lambda_i)^{-k}Q(x)Q(y)\d x\d y\\
        =&\frac12\int\int_{x,y\ge 0}Q(x)Q(y)\left[
        (x+\lambda_q)^{-3}(x+\lambda_i)^{-k}+(y+\lambda_q)^{-3}(y+\lambda_i)^{-k}
        \right]\d x\d y,
    \end{aligned}
    \]
    and similarly,
    \[
    \begin{aligned}
        J_{i,N+k,\mathcal{I}}(\lambda)J_{i,N,\mathcal{I}}^q(\lambda)=\frac12\int\int_{x,y\ge 0}Q(x)Q(y)\left[
        (y+\lambda_q)^{-3}(x+\lambda_i)^{-k}+(x+\lambda_q)^{-3}(y+\lambda_i)^{-k}
        \right]\d x\d y,
    \end{aligned}
    \]
    then we have $J_{i,N+k,\mathcal{I}}^q(\lambda)J_{i,N,\mathcal{I}}(\lambda)-J_{i,N+k,\mathcal{I}}(\lambda)J_{i,N,\mathcal{I}}^q(\lambda)=$
    \[
    \begin{aligned}
        &\frac12\int\int_{x,y\ge 0}Q(x)Q(y)\left[
        (x+\lambda_q)^{-3}(x+\lambda_i)^{-k}+(y+\lambda_q)^{-3}(y+\lambda_i)^{-k}
        \right.\\
        &\left.-(y+\lambda_q)^{-3}(x+\lambda_i)^{-k}-(x+\lambda_q)^{-3}(y+\lambda_i)^{-k}
        \right]\d x\d y \\
        =&\frac12\int\int_{x,y\ge 0}Q(x)Q(y)\left[(x+\lambda_q)^{-3}-(y+\lambda_q)^{-3}\right]\left[(x+\lambda_i)^{-k}-(y+\lambda_i)^{-k}\right]\d x\d y\ge 0,
    \end{aligned}
    \]
    which implies that $\frac{J_{i,N+k,\mathcal{I}}(\lambda)}{J_{i,N,\mathcal{I}}(\lambda)}$ increases with $\lambda_q\ge 0$. 
    % The case for $q\in\mathcal{I}$ can be shown by the same argument.
\end{proof}
\subsection{Proof for Lemma \ref{hnk}}\label{proofhnk}
We divide the proof into two parts. Recall that for all $\mu\ge0$, $K, M\ge 1$, $ N\ge 3$ and $\mu_i\in\R$ for all $1\leq i\leq M$, we defined
    \[
    H_{N}=\int_0^{+\infty} (x+\mu)^{-N}e^{-\frac{K}{(x+\mu)^2}}\prod_{i=1}^M\left(1-F(x+\mu_i)\right) \d x.
    \]
\begin{Lem}\label{newbound13}
    For all $k\in\mathcal{N}^+$, we have
    \[
    \frac{H_{N+k}}{H_N}\leq C_{N,k}\left(\left(\frac{M}{K}\right)^{k/3}\wedge \mu^{-k}\right),
    \]
    where $C_{N,k}$ is a positive constant only depending on $N$ and $k$.
    \begin{proof}
        The upper bound of $\mu^{-k}$ is obvious because $(x+\mu)^{-N-k}\leq \mu^{-k}(x+\mu)^{-N}$ and hence, in the following we assume that $\mu^{-1}\ge C_{N}'\left(\frac{M}{K}\right)^{1/3}$, where $C_{N}'=8\sqrt{N-2}$. Let $u=\frac{1}{x+\mu}$ and 
        \[
        g(u)=u^{N-2}e^{-Ku^2}\prod^M_{i=1}\left(1-F\left(\frac{1}{u}+\mu_i-\mu \right)\right)/\Lambda, u\ge 0,
        \]
        where $\Lambda$ is a constant such that $\int_0^{\mu^{-1}} g(u) \d u=1$. Consider a random variable $U$ with pdf $g(u)$, then clearly, 
        \[
        \frac{H_{N+k}}{H_N}=\E[U^k].
        \]

        Let $\ell(u)=\log g(u)$ and $y_i=\frac{1}{u}+\mu_i-\mu$ for all $1\leq i\leq M$, then when $u\ge 0$, 
        \[
        \frac{\d \ell}{\d u}(u)=\frac{N-2}{u}-2Ku+2\sum_{i=1}^M\frac{\1_{\{y_i\ge 0\}}}{u^2y_i^3\left(e^{\frac{1}{y_i^2}}-1\right)}.
        \]
        Then it suffices to show that when $u\ge u_0:=C_{N}'\left(\frac{M}{K}\right)^{1/3}/2$, $\frac{\d \ell}{\d u}(u)\leq -Ku/2$. Because note that $u_0\ge4K^{-1/3}\ge4K^{-1/2}$ and $\mu^{-1}\ge 2u_0$, then by Lemma \ref{2mu}, we have
        \[
        \E[U^k]\leq u_0^k+\E[U^k\1_{\{\mu^{-1}\ge U\ge u_0\}}]\leq (1+2k!!)u_0^k.
        \]
        Note that $\sup_{x\ge0}\frac{1}{x^3\left(e^{\frac{1}{x^2}}-1\right)}<1$, then we have
        \[
        \frac{\d \ell}{\d u}(u)\leq\frac{N-2}{u}-2Ku+\frac{2M}{u^2}.
        \]
        When $u\ge u_0$, clearly, since $C_N$ is large enough, we have
        \[
        \frac{2M}{u^2}\leq \frac{Ku}{2}.
        \]
        Also, since $u_0>\sqrt{\frac{N-2}{K}}$, when $u\ge u_0$, we have
        $\frac{N-2}{u}\leq Ku$. Therefore, when $u\ge u_0$, $\frac{\d \ell}{\d u}(u)\leq -Ku/2$, which completes our proof. 

    \end{proof}
\end{Lem}
\begin{Lem}\label{newboundlog}
    If $K\ge M$, then we have
    \[
    \frac{H_{4}}{H_3}\leq C\left(\left(\frac{M}{K}\log\left(\frac{K}{M}+1\right)\right)^{1/2}\wedge \mu^{-1}\right),
    \]
    where $C$ is a positive constant.
    \begin{proof}
        We still use the definition of $g$ and $U$ in the proof of Lemma \ref{newbound13} ($N=3$ and $k=1$). Let $u_0=\left(\frac{M}{K}\log\left(\frac{K}{M}+1\right)\right)^{1/2}$. When $K\leq 32M$, by Lemma \ref{newbound13}, the result holds clearly when $C$ is large enough. Hence we then assume that $K>32M$. Then $u_0<1/3$. Similarly, in the following, we also assume that $\mu^{-1}\ge C'u_0$, where $C'>3$ is a large constant to be chosen and is not depending on $K$ and $M$.
        
        For all $s\ge t\ge 1$, we have
        \[
        \frac{g(su_0)}{g(tu_0)}=\frac{s}{t}\mathrm{e}^{-Ku_0^2(s^2-t^2)}\prod_{i=1}^M\frac{1-F\left(\frac{1}{su_0}+\mu_i-\mu\right)}{1-F\left(\frac{1}{tu_0}+\mu_i-\mu\right)}.
        \]
        If $1\leq t\leq 3$, then $tu_0<1$. Hence, by Lemma \ref{ratio1-f}, we have\begin{equation}\label{gsgt}
            \frac{g(su_0)}{g(tu_0)}\leq \frac{s}{t}\mathrm{e}^{-Ku_0^2(s^2-t^2)}\left(\frac{8}{t^2u_0^2}\right)^M\leq s\mathrm{e}^{-Ku_0^2(s^2-t^2)}\left(\frac{8}{u_0^2}\right)^M
            % \leq s\mathrm{e}^{-Ku_0^2(s^2-4)}\left(\frac{8}{u_0}\right)^M
            .
        \end{equation}
        Then on the one hand, for all $t\in[1,2]$, we have
        \[
         \frac{g(3u_0)}{g(tu_0)}\leq 3\mathrm{e}^{-5Ku_0^2}\left(\frac{8}{u_0^2}\right)^M\leq 3\mathrm{e}^{-5M\log\left(\frac{K}{M}+1\right)}\left(\frac{8K}{M}\right)^M,
        \]
        where we used the definition of $u_0$ in the last inequality. Since $\frac{K}{M}>8$, we have
        \[
        \frac{g(3u_0)}{g(tu_0)}\leq 3\mathrm{e}^{-5M\log\left(\frac{K}{M}+1\right)+2M\log\left(\frac{K}{M}\right)}<1.
        \]
        Then since $C'\ge 2$,
        \[
        1\ge \int_{u_0}^{\mu^{-1}}g(u) \d u\ge \int_{u_0}^{2u_0}g(u) \d u\ge u_0 g(3u_0),
        \]
        which implies that $g(3u_0)\leq u_0^{-1}.$

        On the other hand, by \eq{gsgt}, for all $s\ge 3$, we have
        \[
        \frac{g(su_0)}{g(3u_0)}\leq s\mathrm{e}^{-Ku_0^2(s^2-9)}\left(\frac{8}{u_0^2}\right)^M\leq\mathrm{e}^{\log (s)-M\log\left(\frac{K}{M}+1\right)(s^2-9)+M\log(8)+M\log\left(\frac{K}{M}\right)}.
        \]
        Then since $K>32M\ge 32$, one can find $C''>3$ that is not depending on $K$ and $M$ and large enough (one can then pick an $C'$ larger than $C''$) such that for all $s\ge C''$, we have
        \[
        \frac{g(su_0)}{g(3u_0)}\leq\mathrm{e}^{-\frac{M}{2}\log\left(\frac{K}{M}+1\right)s^2}.
        \]
        Therefore,
        \[
        \begin{aligned}
            \int_{C''}^{\frac{\mu^{-1}}{u_0}}s g(su_0)\d s\leq &g(3u_0)\int_{C''}^{+\infty}s\mathrm{e}^{-M\log\left(\frac{K}{M}+1\right)s^2/2} \d s\leq u_0^{-1}\int_{C''}^{+\infty}s\mathrm{e}^{-M\log\left(\frac{K}{M}+1\right)s^2/2} \d s\\
            =&u_0^{-1}\left(M\log\left(\frac{K}{M}+1\right)\right)^{-1}\mathrm{e}^{-C''^2M\log\left(\frac{K}{M}+1\right)/2}\leq u_0^{-1}.
        \end{aligned}
        \]
        Then we have
        \[
        \begin{aligned}
            \E[U]\leq C''u_0+\int_{C''u_0}^{\mu^{-1}} ug(u)\d u=C''u_0+u_0^2\int_{C''}^{\frac{\mu^{-1}}{u_0}} sg(su_0)\d s\leq (C''+1)u_0.
        \end{aligned}
        \]
    \end{proof}
\end{Lem}
\subsection{Lower Bounds}\label{lowertight}
In this section, we will prove lower bounds for Lemma \ref{hnk}, showing that the logarithmic term and $M^{1/3}$ are inevitable.
\begin{Lem}\label{log}
    For all $\mu\in\R$, $K\ge 1$, and $ N\ge 3$, define
    \[
    U_{N}(\mu)=\int_0^{+\infty} x^{-N}e^{-\frac{K}{x^2}}\left(1-F(x+\mu)\right) \d x.
    \]
    Then there exists $C>0$ such that for all $K\ge 2$, we have
    \[
    \sup_{\mu\in\R}\frac{U_{4}(\mu)}{U_3(\mu)}\ge C\sqrt{\frac{\log(K)}{K}}.
    \] 
    \begin{proof}
        Let $w=e^{-\frac{K}{x^2}}$, then $x(w)=\sqrt{\frac{K}{-\log w}}$ and
        \[
        U_{N}(\mu)=\frac12 K^{-\frac{N-1}{2}}\int_{0}^1 (-\log w)^{\frac{N-3}{2}}(1-F(x(w)+\mu))\d w,
        \]
        which implies that
        \[
        \sqrt{K}\frac{U_{4}(\mu)}{U_3(\mu)}=\frac{\int_{0}^1 (-\log w)^{\frac{1}{2}}(1-F(x(w)+\mu))\d w}{\int_{0}^1 [1-F(x(w)+\mu)]\d w}=\E[(-\log(W))^{1/2}],
        \]
        where $W$ is a random variable with p.d.f. proportional to $(1-F(x(w)+\mu))\1_{\{w\in[0,1]\}}$. Clearly, for all $s\ge 0$, we have
        \begin{equation}\label{expe}
            \E[(-\log(W))^{1/2}]\ge \sqrt{s}\P(W\leq e^{-s}).
        \end{equation}
        Let $\mu=-\sqrt{\frac{2K}{\log(K)}}$ and $s=\frac{K}{\mu^2}=\frac{\log(K)}{2}$. Then when $w\leq e^{-s}$, one can see that $x(w)+\mu\leq 0$, which implies that 
        \[
        \int_{0}^{e^{-s}} \big [1-F(x(w)+\mu)\big] \d w=e^{-s}=K^{-1/2}.
        \]
        Then it suffices to show that
        \begin{equation}\label{small o}
        \int_{e^{-s}}^1  \big[ 1-F(x(w)+\mu)\big] \d w=e^{-s}=\mathcal{O}(K^{-1/2}),
        \end{equation}
        since then we have
        \[
        \P(W\leq e^{-s})=\frac{K^{-1/2}}{K^{-1/2}+\mathcal{O}(K^{-1/2})}\ge C,
        \]
        where $C$ is a positive constant. Combining \eq{expe} and \eq{small o} leads to the desired result.

        To show \eq{small o}, one should note that when $w\ge e^{-(K^{-1/4}+s^{-1/2})^{-2}}$,
        \[
        1-F(x(w)+\mu)=1-e^{-\frac{1}{(x(w)+\mu)^2}}\leq \frac{1}{(x(w)+\mu)^2}=\frac{1}{K\left(
        \sqrt{\frac{1}{-\log w}}-\sqrt{\frac{1}{s}}
        \right)^2}\leq K^{-1/2},
        \]
        where we used that $1-e^{-x}\leq x$ for all $x\in\R$. Then
        \[
        \int_{e^{-(K^{-1/4}+s^{-1/2})^{-2}}}^1 \big[1-F(x(w)+\mu) \big]\d w\leq K^{-1/2}.
        \]
        Finally, since
        \[
        e^{-(K^{-1/4}+s^{-1/2})^{-2}}-e^{-s}=e^{-s}\cdot\left(e^{s-s(K^{-1/4}s^{1/2}+1)^{-2}}-1\right),
        \]
        and noting that, by $s=\frac{\log(K)}{2}$,
        \[
        s-s(K^{-1/4}s^{1/2}+1)^{-2}=\frac{s(K^{-1/2}s+2K^{-1/4}s^{1/2})}{(K^{-1/4}s^{1/2}+1)^{2}}=o(1),
        \]
        we then have
        \[
        \int_{e^{-s}}^{e^{-(K^{-1/4}+s^{-1/2})^{-2}}} \big[ 1-F(x(w)+\mu) \big] \d w\leq e^{-(K^{-1/4}+s^{-1/2})^{-2}}-e^{-s}=o(e^{-s})=o(K^{-1/2}),
        \]
        where we used that $e^{o(1)}-1=o(1).$ It suffices to put everything together.
    \end{proof}
\end{Lem}
\begin{Lem}
    For all $\mu\in\R$, $M, K\ge 1$, and $ N\ge 3$, define
    \[
    R_{N}(\mu)=\int_0^{+\infty} x^{-N}e^{-\frac{K}{x^2}}\left(1-F(x+\mu)\right)^M \d x.
    \]
    Then there exists $C>0$ such that for all $M\ge 2K$, we have
    \[
    \sup_{\mu\in \R}\frac{R_{4}(\mu)}{R_3(\mu)}\ge C\left(\frac{M}{K}\right)^{1/3}.
    \] 
    \begin{proof}
        % We follow the proof of Lemma \ref{newbound}. 
        We still use the definition of $g$ and $U$ in the proof of Lemma \ref{newbound13} ($N=3$ and $k=1$).
        Our plan is to show that that exists $\mu\in\R$ and corresponding $C>0$ such that when $0\leq u\leq u_0:=C\left(\frac{M}{K}\right)^{1/3}$, $\frac{\d \ell}{\d u}(u)\ge 0$, which implies that $g(u)$ increases with $u$ when $u\leq u_0$. Then we have
        \[
        \P\left(U\leq \frac{u_0}{2}\right)=\int_{0}^{\frac{u_0}{2}}g(u)\d u\leq \int_{\frac{u_0}{2}}^{u_0}g(u)\d u,
        \]
        which implies that $\P\left(U\leq \frac{u_0}{2}\right)\leq \frac{1}{2}.$ Then by Markov's inequality, we have
        \[
        \E[U]\ge \frac{u_0}{2}\P\left(U> \frac{u_0}{2}\right)\ge\frac{u_0}{4},
        \]
        which is just the desired result.

        To this end, let $\mu=1>0$ and $y=\frac{1}{u}+1$, then
        \[
        \frac{\d \ell}{\d u}(u)=\frac{1}{u}-2Ku+\frac{2M}{u^2y^3\left(e^{\frac{1}{y^2}}-1\right)}=\frac{1}{u}+2Ku\left(
        \frac{M/K}{u^3y^3\left(e^{\frac{1}{y^2}}-1\right)}-1
        \right).
        \]
        Since $1\leq y$, when $u\leq \left(\frac{M}{K}\right)^{\frac13}(e-1)^{-\frac13}-1$, we have
        \[
        \frac{M/K}{u^3y^3\left(e^{\frac{1}{y^2}}-1\right)}=\frac{M/K}{(u+1)^3\left(e^{\frac{1}{y^2}}-1\right)}\ge\frac{M/K}{(u+1)^3\left(e-1\right)}\ge 1.
        \]
        Take $C=(e-1)^{-\frac13}-2^{-\frac13}>0$ and then one can see that when $M\ge 2K$, $u_0:=C\left(\frac{M}{K}\right)^{1/3}\leq \left(\frac{M}{K}\right)^{\frac13}(e-1)^{-\frac13}-1$. Then when $0\leq u\leq u_0:=C\left(\frac{M}{K}\right)^{1/3}$, $\frac{\d \ell}{\d u}(u)\ge 0$.
    \end{proof}
\end{Lem}

\section{Stability Term}\label{stab sec}
In this section, we provide our results related to the stability term. We start with showing some important properties for $V_{i,N}$.
\begin{Lem}\label{lowerlimit}
    For all $\mathcal{I}\subset\{1,\cdots,d\}$ such that $|\mathcal{I}|<m$ and $i\notin\mathcal{I}$, where $1\leq i\leq d$, we have
    \[
    V_{i,N}^{\mathcal{I}}(\lambda)=\int_{-\min\limits_{j\notin \mathcal{I}}\lambda_j}^{\infty} \frac{1}{(x+\lambda_i)^N}   \prod_{q \in \mathcal{I}} (1 - F(x+\lambda_q)) \prod_{q \notin \mathcal{I}} F(x+\lambda_q)    \d x.
    \]
    Then for all $1\leq i\leq d$ such that $\sigma_i(\lambda)>m$, we have
    \[
    V_{i,N}(\lambda)=\int_{-\max\limits_{\mathcal{I},|\mathcal{I}|<m}\min\limits_{j\notin \mathcal{I}}\lambda_j}^{\infty} \frac{1}{(x+\lambda_i)^N} e^{-1/(x+\lambda_i)^2} \sum_{s=0}^{m-1} \sum_{\mathcal{I} \subseteq  \{1,\dots,d\} \setminus \{i\}, |\mathcal{\mathcal{I}}|=s} \left[ \prod_{q \in \mathcal{I}} (1 - F(x+\lambda_q)) \prod_{q \notin \mathcal{I}, q \neq i} F(x+\lambda_q) \right] \d x.
    \]
    \begin{proof}
        For the first result, it suffices to note that $F(x)=0$ when $x\leq 0$, then when $x\leq -\min\limits_{j\notin \mathcal{I}}\lambda_j$, the integrand in \eq{vini} is just $0$. For the second result, recall that 
        \[V_{i,N}(\lambda)=\sum_{s=0}^{m-1} \sum_{\mathcal{I} \subseteq  \{1,\dots,d\} \setminus \{i\}, |\mathcal{\mathcal{I}}|=s}V_{i,N}^{\mathcal{I}}(\lambda)\]
        and note that for all $\mathcal{I}\subset\{1,\cdots,d\}$ such that $|\mathcal{I}|<m$ and $i\notin\mathcal{I}$,
        \[
        -\max\limits_{\mathcal{I},|\mathcal{I}|<m}\min\limits_{j\notin \mathcal{I}}\lambda_j=\min\limits_{\mathcal{I},|\mathcal{I}|<m}\left(-\min\limits_{j\notin \mathcal{I}}\lambda_j\right)\leq-\min\limits_{j\notin \mathcal{I}}\lambda_j, 
        \]
        then the result follows from that for every $V_{i,N}^{\mathcal{I}}(\lambda)$, the lower limit of the integral can be further reduced to \(-\max\limits_{\mathcal{I},|\mathcal{I}|<m}\min\limits_{j\notin \mathcal{I}}\lambda_j\).
    \end{proof}
\end{Lem}

\begin{Lem}\label{monontone1}
The followings hold:
\begin{enumerate}
    %\item $V_{i,n}(\lambda)\ge 0.$
    \item For all $N\ge2$, $V_{i,N}(\lambda)\leq \frac{\underline{\lambda}_i^{1-N}}{N-1}.$
    %\item For all $t\in \R^d$, $V_{i,n}(\lambda+t\mathbf{1}_d)=V_{i,n}(\lambda)$. 
    \item For all $1\leq i\leq d$ and $N\ge 3$, $V_{i,N}(\lambda)$ is increasing in all $\lambda_j, j\ne i$ and decreasing in $\lambda_i$.
    \item $\bar{V}_{i,N}(\lambda):=\frac{\Gamma(\frac{N-1}{2})}{2}-V_{i,N}(\lambda)=$
    \[
    \int_{-\lambda_i}^{\infty} \frac{1}{(x+\lambda_i)^N} e^{-1/(x+\lambda_i)^2} \sum_{s=m}^{d-1} \sum_{\mathcal{I} \subseteq  \{1,\dots,d\} \setminus \{i\}, |\mathcal{\mathcal{I}}|=s} \left[ \prod_{q \in \mathcal{I}} (1 - F(x+\lambda_q)) \prod_{q \notin \mathcal{I}, q \neq i} F(x+\lambda_q) \right] \d x\ge 0.
    \]
\end{enumerate}
\begin{proof}
    For the first result, obviously, it suffices to consider the case when $\sigma_i(\lambda)>m.$  By Lemma \ref{lowerlimit}, we have
\[
\begin{aligned}
    V_{i,N}(\lambda)=&\int_{-\max\limits_{\mathcal{I},|\mathcal{I}|<m}\min\limits_{j\notin \mathcal{I}}\lambda_j}^{\infty} \frac{1}{(x+\lambda_i)^N} e^{-1/(x+\lambda_i)^2} \sum_{s=0}^{m-1} \sum_{\mathcal{I} \subseteq  \{1,\dots,d\} \setminus \{i\}, |\mathcal{\mathcal{I}}|=s} \left[ \prod_{q \in \mathcal{I}} (1 - F(x+\lambda_q)) \prod_{q \notin \mathcal{I}, q \neq i} F(x+\lambda_q) \right] \d x\\
    \leq &\int_{-\max\limits_{\mathcal{I},|\mathcal{I}|<m}\min\limits_{j\notin \mathcal{I}}\lambda_j}^{\infty} \frac{1}{(x+\lambda_i)^N}\d x=\int_{0}^{\infty} \frac{1}{(z+\underline{\lambda}_i)^N}\d z=\frac{\underline{\lambda}_i^{1-N}}{N-1},
\end{aligned}
\]
where in the inequality, we upper bound the conditional probability inside (see Lemma \ref{phii=2v}) by $1$.

    For the rest results, consider a random variable $r_i'$ with density
    \[
    g(x)=\frac{x^{-N}e^{-1/x^2}\1_{\{x>0\}}}{\int_0^{+\infty} x^{-N}e^{-1/x^2}\d x}=\frac{2}{\Gamma(\frac{N-1}{2})}x^{-N}e^{-1/x^2}\1_{\{x>0\}}.
    \]
    Then similar to \eq{defi phi}, 
    \[
    \begin{aligned}
        V_{i,N}(\lambda)=\frac{\Gamma(\frac{N-1}{2})}{2}\P(&\text{there exist at most $m-1$ of $r_1-\lambda_1,\cdots,r_{i-1}-\lambda_{i-1}, r_{i+1}-\lambda_{i+1},\cdots,$} \\
        &\text{$r_d-\lambda_d$ that are larger than $r'_i-\lambda_i$}),
    \end{aligned}
    \]
    where $r_1,\cdots,r_{i-1},r_{i+1},\cdots,r_d\iidsim\mathcal{F}_2.$ Then these properties hold obviously.
\end{proof}
\end{Lem}

\begin{Lem}\label{-k/2}
    For all $1\leq i\leq d$, we have
    \[
    \frac{V_{i,4}(\lambda)}{V_{i,3}(\lambda)}\leq \underline{\lambda}_{i}^{-1}\wedge C\begin{cases}
        \sqrt{\frac{m\log(d)}{\sigma_i(\lambda)-m}} & \sigma_i(\lambda)>2m
        \\
        %\sqrt{\log(m)+1} 
        m^{1/3}& 
        \sigma_i(\lambda)\leq 2m,\end{cases}
    \]
    where $C$ is a positive constant.
    \begin{proof}
    % It suffices to show the first result and then the second can be derived by the same way. 
    For any $\mathcal{I}\subseteq \{1,\cdots,d\}$ such that $|\mathcal{I}|<m$ and $i\notin\mathcal{I}$, by Lemma \ref{lowerlimit}, we have
    \begin{equation}\label{change viin}
                     \begin{aligned}
          V_{i,N}^{\mathcal{I}}(\lambda)=&\int_{-\min\limits_{j\notin \mathcal{I}}\lambda_j}^{\infty} \frac{1}{(x+\lambda_i)^N}   \prod_{q \in \mathcal{I}} (1 - F(x+\lambda_q)) \prod_{q \notin \mathcal{I}} F(x+\lambda_q)    \d x\\
          =&\int_{0}^{\infty} \frac{1}{(z+\underline{\lambda}^\mathcal{I}_i)^N}   \prod_{q \in \mathcal{I}} (1 - F(z+\underline{\lambda}^\mathcal{I}_q)) \prod_{q \notin \mathcal{I}} F(z+\underline{\lambda}^\mathcal{I}_q)    \d z,
         \end{aligned}
        \end{equation}
        where we denoted that $\underline{\lambda}^\mathcal{I}_q:=\lambda_q-\min\limits_{j\notin \mathcal{I}}\lambda_j$ for all $1\leq q\leq d$ and we denoted that $z=x+\min\limits_{j\notin \mathcal{I}}\lambda_j.$ Then, clearly, 
        $
        \frac{V^{\mathcal{I}}_{i,4}(\lambda)}{V^{\mathcal{I}}_{i,3}(\lambda)}\leq(\underline{\lambda}_{i}^{\mathcal{I}})^{-1},
        $ which implies that
        \[
        \frac{V_{i,4}(\lambda)}{V_{i,3}(\lambda)}\leq\max\limits_{\mathcal{I},|\mathcal{I}|<m,i\notin\mathcal{I}}(\underline{\lambda}_{i}^{\mathcal{I}})^{-1}=(\lambda_i-\max\limits_{\mathcal{I},|\mathcal{I}|<m,i\notin\mathcal{I}}\min_{j\notin\mathcal{I}}\lambda_j)^{-1}\leq\underline{\lambda}_{i}^{-1}.
        \]
        %If $\sigma_i(\lambda)>m$, 
        For the second part, by \eq{change viin} and Lemma \ref{monontone2}, we have
        \begin{equation}\label{v j i}
                    \frac{V_{i,4}^{\mathcal{I}}(\lambda)}{V^{\mathcal{I}}_{i,3}(\lambda)}=\frac{J_{i,4,\mathcal{I}}(\underline{\lambda}^\mathcal{I})}{J_{i,N,\mathcal{I}}(\underline{\lambda}^\mathcal{I})}\leq \frac{J_{i,4,\mathcal{I}}(\lambda^\star)}{J_{i,N,\mathcal{I}}(\lambda^\star)}=\frac{\int_{0}^{\infty} \frac{1}{(z+\underline{\lambda}^\mathcal{I}_i)^{4}}     e^{-\frac{\sigma'_i(\lambda)}{(z+\underline{\lambda}^\mathcal{I}_i)^2}}\prod_{q\in\mathcal{I}}(1 - F(z+\lambda_q^\star))\d z}{\int_{0}^{\infty} \frac{1}{(z+\underline{\lambda}^\mathcal{I}_i)^3}    e^{-\frac{\sigma'_i(\lambda)}{(z+\underline{\lambda}^\mathcal{I}_i)^2}}\prod_{q\in\mathcal{I}}(1 - F(z+\lambda_q^\star))\d z},
        \end{equation}
        where we denoted that
        \[
        \lambda_q^
        \star= \begin{cases}
        +\infty & q\notin\mathcal{I} \text{ and } \lambda_q\ge\lambda_i
        \\
        \underline{\lambda}^\mathcal{I}_i & q\notin\mathcal{I} \text{ and } \lambda_q\leq\lambda_i 
        \\
        %  0 & q\in\mathcal{I} \text{ and }\lambda_q\ge\lambda_i
        % \\
        \underline{\lambda}_q^{\mathcal{I}} & 
        q\in\mathcal{I},\end{cases}
        \]
        and $\lambda_i$ is the $\sigma'_i(\lambda)$-th smallest in $\{\lambda_q\}_{q\notin\mathcal{I}}$. To apply Lemma \ref{hnk}, one should note that here $K=\sigma'_i(\lambda)$ and $M=|\mathcal{I}|$. If $\sigma_i(\lambda)>2m$, then $K\ge \sigma_i(\lambda)-m\ge m\ge M$, then by the second result in Lemma \ref{hnk}, 
        there exists $C>0$ such that the right hand in \eq{v j i}
        \[
        % =\frac{I_{4,\sigma'_i(\lambda)}(\underline{\lambda}^\mathcal{I}_i)}{I_{3,\sigma'_i(\lambda)}(\underline{\lambda}^\mathcal{I}_i)}\leq C_{3,k}((\underline{\lambda}^\mathcal{I}_i)^2\vee \sigma'_i(\lambda))^{-\frac{k}{2}}
        % \leq C_{N,k}(\underline{\lambda}_i^2\vee \widetilde{\sigma}_i(\lambda))^{-\frac{k}{2}},
        %\leq C_{N,k}\sigma'_i(\lambda)^{-k/2}\leq C_{N,k}\widetilde{\sigma}_i(\lambda)^{-k/2},
        \leq C\sqrt{\frac{M}{K}\log\left(\frac{K}{M}+1\right)}
        \leq C\sqrt{\frac{m\log(d)}{\sigma_i(\lambda)-m}}.
        \]

        If $\sigma_i(\lambda)\leq 2m$, similarly, by the first result in Lemma \ref{hnk}, the right hand in \eq{v j i} is less than
        \[
            % \frac{V_{i,4}^{\mathcal{I}}(\lambda)}{V^{\mathcal{I}}_{i,3}(\lambda)}\leq \frac{\int_{0}^{\infty} \frac{1}{(z+\underline{\lambda}^\mathcal{I}_i)^{4}}     e^{-\frac{1}{(z+\underline{\lambda}^\mathcal{I}_i)^2}}\prod_{q\in\mathcal{I}}(1 - F(z+\lambda_q^\star))\d z}{\int_{0}^{\infty} \frac{1}{(z+\underline{\lambda}^\mathcal{I}_i)^3}    e^{-\frac{1}{(z+\underline{\lambda}^\mathcal{I}_i)^2}}\prod_{q\in\mathcal{I}}(1 - F(z+\lambda_q^\star))\d z},
            C\left(\frac{M}{K}\right)^{\frac{1}{3}}\leq m^{1/3}.
        \]
        % because one can simply let $\lambda_q$ increase to $+\infty$ for every $q\not\in\mathcal{I}$ and $q\ne i.$ Then by Lemma \ref{hnk}, one can see that it's less than $C\sqrt{\log(m)+1}$.
    \end{proof}
\end{Lem}

\begin{Lem}\label{vbar}
    For all $1\leq i\leq d$, $N\ge 3$ and $k\ge 1$, we have
    \[
    \frac{\bar{V}_{i,N+k}(\lambda)}{\bar{V}_{i,N}(\lambda)}\leq C_{N,k}d^{\frac{k}{3}},
    \]
    where $C_{N,k}$ is a positive constant.
    \begin{proof}
        Recall that in Lemma \ref{monontone1} we have
        \[
    \int_{-\lambda_i}^{\infty} \frac{1}{(x+\lambda_i)^N} e^{-1/(x+\lambda_i)^2} \sum_{s=m}^{d-1} \sum_{\mathcal{I} \subseteq  \{1,\dots,d\} \setminus \{i\}, |\mathcal{\mathcal{I}}|=s} \left[ \prod_{q \in \mathcal{I}} (1 - F(x+\lambda_q)) \prod_{q \notin \mathcal{I}, q \neq i} F(x+\lambda_q) \right] \d x\ge 0,
    \]
        then the proof follows from the case $\sigma_{i}(\lambda)\leq 2m$ in Lemma \ref{-k/2} by noting that $|\mathcal{I}|\leq d$. 
    \end{proof}
\end{Lem}

\begin{Lem}\label{lambda xi}
    There exists $C>0$ such that for all $a\in[0,d^{-1}]$ and $1\leq i\leq d$,
    \[
    \frac{ \bar{V}_{i,6}(\lambda+ae_i)}{ \bar{V}_{i,6}(\lambda)}\leq C.
    \]
    \begin{proof}
    It suffices to show that $\frac{\pd}{\pd\lambda_i}\log (\bar{V}_{i,6}(\lambda))$ is upper bounded by $Cd$. By the definition,
\[
\frac{\pd}{\pd\lambda_i}\log (\bar{V}_{i,6}(\lambda))=\frac{\frac{\pd}{\pd\lambda_i}\bar{V}_{i,6}(\lambda)}{\bar{V}_{i,6}(\lambda)},
\]
where
\[
\frac{\pd}{\pd\lambda_i}\bar{V}_{i,6}(\lambda)=\int_{-\lambda_i}^{\infty} \frac{\pd}{\pd\lambda_i}\left(\frac{1}{(x+\lambda_i)^6} e^{-1/(x+\lambda_i)^2}\right) \sum_{s=m}^{d-1} \sum_{\mathcal{I} \subseteq  \{1,\dots,d\} \setminus \{i\}, |\mathcal{\mathcal{I}}|=s}  \left[ \prod_{q \in \mathcal{I}} (1 - F(x+\lambda_q)) \prod_{q \notin \mathcal{I}, q \neq i} F(x+\lambda_q) \right] \d x,
\]
since $\lim\limits_{x\to-\lambda_i}\frac{1}{(x+\lambda_i)^6} e^{-1/(x+\lambda_i)^2}=0$
. Note that $\frac{\pd}{\pd\lambda_i}\left(\frac{1}{(x+\lambda_i)^6} e^{-1/(x+\lambda_i)^2}\right)\leq \frac{2}{(x+\lambda_i)^{9}} e^{-1/(x+\lambda_i)^2}$, then it's clear that $\frac{\pd}{\pd\lambda_i}\bar{V}_{i,6}(\lambda)\leq 2\bar{V}_{i,9}(\lambda).$ Hence, by Lemma \ref{vbar}, there exits $C>0$ such that $\frac{\pd}{\pd\lambda_i}\log (\bar{V}_{i,6}(\lambda))\leq Cd.$
    \end{proof}
\end{Lem}
Then we show our the results about the continuity of $\phi$.
\begin{Lem}\label{w1-w2}
    There exists $C>0$ such that for all $1\leq i\leq d$, $a>0$ and $\lambda\in\R^d$, if $w=\phi_i(\lambda)$ and $w'=\phi_i(\lambda+ae_i)$, then the followings hold:
    \begin{enumerate}
        \item $w-w'\leq Cwa\cdot \underline{\lambda}_{i}^{-1}\wedge \begin{cases}
        \sqrt{\frac{m\log(d)}{\sigma_i(\lambda)-m}} & \sigma_i(\lambda)>2m
        \\
        m^{1/3}%\sqrt{\log(m)+1} 
        & 
        \sigma_i(\lambda)\leq 2m.\end{cases}$
        \item $w-w'\leq Cd (1-w)a,$ if $a\leq d^{-1}.$
    \end{enumerate}
    \begin{proof}
        For the first result, note that for all $t\in[0,1]$
        \[
        \frac{\d}{\d t}\phi_i(\lambda+(1-t)a e_i)=-a\frac{\pd\phi_i}{\pd\lambda_i}(\lambda+(1-t)ae_i).
        \]
        Then,  
        \[
        -\frac{\pd\phi_i}{\pd\lambda_i}(\lambda+(1-t)ae_i)=6V_{i,4}(\lambda+(1-t)ae_i)-4V_{i,6}(\lambda+(1-t)ae_i)\leq 6V_{i,4}(\lambda+(1-t)ae_i)\leq 6V_{i,4}(\lambda),
        \]
        where we used Lemma \ref{monontone1} in the final inequality. Recall that $w=\phi_i(\lambda)=2V_{i,3}(\lambda)$, then by Lemma \ref{-k/2}, 
        \[
        -\frac{\pd\phi_i}{\pd\lambda_i}(\lambda+(1-t)ae_i)\leq C'w\cdot \underline{\lambda}_{i}^{-1}\wedge \begin{cases}
        \sqrt{\frac{m\log(d)}{\sigma_i(\lambda)-m}} & \sigma_i(\lambda)>2m
        \\
        m^{1/3}%\sqrt{\log(m)+1}
        & 
        \sigma_i(\lambda)\leq 2m.\end{cases}
        \]
        Therefore,
        \begin{equation}\label{w1w2}
             w-w'=\int_0^1 \frac{\d}{\d t}\phi_i(\lambda+(1-t)ae_i)\d t\leq C'wa\cdot \underline{\lambda}_{i}^{-1}\wedge \begin{cases}
        \sqrt{\frac{m\log(d)}{\sigma_i(\lambda)-m}} & \sigma_i(\lambda)>2m
        \\
        m^{1/3}%\sqrt{\log(m)+1}
        & 
        \sigma_i(\lambda)\leq 2m.\end{cases}
        \end{equation}
        
        For the second result, let $\bar{\phi}=\mathbf{1}-\phi$, then $w-w'=\bar{\phi}_i(\lambda+ae_i)-\bar{\phi}_i(\lambda)$. Similarly, for all $t\in[0,1]$,
        \[
        \frac{\d}{\d t}\bar{\phi}_i(\lambda+tae_i)= a\frac{\pd\bar{\phi}_i}{\pd\lambda_i}(\lambda+tae_i).
        \]
        Since now $\bar{\phi}_i(\lambda)=2\bar{V}_{i,3}(\lambda)$, then clearly,
        \[
        \frac{\pd\bar{\phi}_i}{\pd\lambda_i}(\lambda)=-6\bar{V}_{i,4}(\lambda)+4\bar{V}_{i,6}(\lambda)\leq 4\bar{V}_{i,6}(\lambda).
        \]
        Hence, combing Lemma \ref{lambda xi} and Lemma \ref{vbar}, we have
        \[
        \frac{\pd\bar{\phi}_i}{\pd\lambda_i}(\lambda+tae_i)\leq 4\bar{V}_{i,6}(\lambda+tae_i)\leq C\bar{V}_{i,6}(\lambda)\leq C'd\bar{V}_{i,3}(\lambda)=C''d(1-w).
        \]
        Finally, one can obtain the result by the way similar to \eq{w1w2}.
    \end{proof}
\end{Lem}

\subsection{Proof for Lemma \ref{stab1}\label{proofstab1}}
    \begin{proof}
        By Lemma \ref{monontone1}, 
        \[
        \phi_i(\eta_t\hat{L}_{t+1})\ge \phi_i(\eta_t\hat{L}_{t}+\eta_t\hat{\ell}_{t,i}\cdot e_i),
        \]
        then
        \[
        \hat{\ell}_{t,i}\left(\phi_i(\eta_t \hat{L}_t)-\phi_i(\eta_t \hat{L}_{t+1})\right)\leq \hat{\ell}_{t,i}\left(\phi_i(\eta_t \hat{L}_t)-\phi(\eta_t\hat{L}_{t}+\eta_t\hat{\ell}_{t,i}\cdot e_i)\right).
        \]
        Denote that 
        \[
        \Lambda=\underline{\hat{L}}_{t,i}^{-1}\wedge \eta_t\begin{cases}
        \sqrt{\frac{m\log(d)}{\sigma_i(\hat{L}_t)-m}} & \sigma_i(\hat{L}_t)>2m
        \\
        m^{1/3}%\sqrt{\log(m)+1} 
        & 
        \sigma_i(\hat{L}_t)\leq 2m.\end{cases}
        \]
         Hence, by Lemma \ref{w1-w2} 
        \[
        \hat{\ell}_{t,i}(w-w')
        \leq 
        % C\eta_t(\underline{\lambda}_i^2\vee \widetilde{\sigma}_i(\lambda))^{-\frac{1}{2}}
        C\Lambda w\hat{\ell}_{t,i}^2\leq 
        % C\eta_t(\underline{\lambda}_i^2\vee \widetilde{\sigma}_i(\lambda))^{-\frac{1}{2}}
        C\Lambda wK_{t,i}^2A_{t,i},
        \]
        where we denoted that $w=\phi_i(\eta_t \hat{L}_t)$ and $w'=\phi_i(\eta_t \hat{L}_{t}+\eta_t\hat{\ell}_{t,i}\cdot e_i)$. By Lemma \ref{k2ge}, $\E[K_{t,i}^2\mid \mathscr{F}_{t-1},A_t]\leq 2w^{-2}.$ Then
        \[
        \E[\hat{\ell}_{t,i}(w-w')\mid\mathscr{F}_{t-1},A_t]\leq 2
        % C\eta_t(\underline{\lambda}_i^2\vee \widetilde{\sigma}_i(\lambda))^{-\frac{1}{2}}
        C\Lambda
        \cdot w^{-1}A_{t,i}.
        \]
        Hence, since $\E[A_{t,i}\mid\mathscr{F}_{t-1}]=w$, then
        \[
        \E[\hat{\ell}_{t,i}(w-w')\mid\mathscr{F}_{t-1}]\leq 2C\Lambda.
        \]
    \end{proof}

\subsection{Proof for Lemma \ref{stab3}}\label{proofstab3}
\begin{proof}
        By Lemma \ref{stab1}, it's clear that
        \[
        \sum_{i=m+1}^d \E\left[\hat{\ell}_{t,i}\left(\phi_i(\eta_t \hat{L}_t)-\phi_i(\eta_t \hat{L}_{t+1})\right)\mid\mathscr{F}_{t-1}\right]\leq C \sum\limits_{i=m+1}^d \underline{\hat{L}}_{t,i}^{-1},
        \]
        then it suffices to tackle the sum for $1\leq i\leq m$. By Lemma \ref{w1-w2} and following the same argument in Lemma 
        \ref{stab1}, for all $1\leq i\leq m$, we have
        \[
        \E\left[\1_{\{\eta_tK_{t,i}\leq d^{-1}\}}\hat{\ell}_{t,i}\left(\phi_i(\eta_t \hat{L}_t)-\phi_i(\eta_t \hat{L}_{t+1})\right)\mid\mathscr{F}_{t-1}\right]\leq C \eta_t dw_{t,i}^{-1}(1-w_{t,i}),
        \]
        where we denoted that $w_t=\phi(\eta_t\hat{L}_t)$ and we used the fact that $\eta_t\hat{\ell}_{t,i}\leq d^{-1}$ when $\eta_tK_{t,i}\leq d^{-1}$. Note that by Lemma \ref{wstarbound}, when $\sum\limits_{i=m+1}^d (\eta_t\underline{\hat{L}}_{t,i})^{-2}<\frac{1}{2m}$, for all $1\leq i\leq m$,
        \begin{equation}\label{wti12}
            w_{t,i}\ge w_\star\ge1/2.
        \end{equation}
        Hence,
        \[
        \sum_{i=1}^m \E\left[\1_{\{\eta_tK_{t,i}\leq d^{-1}\}}\hat{\ell}_{t,i}\left(\phi_i(\eta_t \hat{L}_t)-\phi_i(\eta_t \hat{L}_{t+1})\right)\mid\mathscr{F}_{t-1}\right]\leq 2C\eta_t d \sum_{i=1}^m(1-w_{t,i})=2C\eta_t d\sum_{i=m+1}^d w_{t,i},
        \]
        where we used that $\sum\limits_{i=1}^d w_{t,i}=m$ by Lemma \ref{sum=m}. 
        Finally, for all $1\leq i\leq m$, 
        \[
        \begin{aligned}
            \E\left[\1_{\{\eta_tK_{t,i}>d^{-1}\}}\hat{\ell}_{t,i}\left(\phi_i(\eta_t \hat{L}_t)-\phi_i(\eta_t \hat{L}_{t+1})\right)\mid\mathscr{F}_{t-1}\right]&\leq \E\left[\1_{\{\eta_tK_{t,i}>d^{-1}\}}A_{t,i}K_{t,i}\mid\mathscr{F}_{t-1}\right]\\&=   \E\left[\1_{\{\eta_tK_{t,i}>d^{-1}\}}w_{t,i}K_{t,i}\mid\mathscr{F}_{t-1}\right] ,
        \end{aligned}
        \]
        which, by Lemma \ref{k2ge}, is less than
        \[
        (1-w_{t,i})^{\lfloor d^{-1}\eta_t^{-1}\rfloor}\leq (1-w_{t,i})^{\frac{1}{2\eta_td}}\leq 2^{-\frac{1}{2\eta_td}},
        \]
        where we used \eq{wti12} in the final inequality. It suffices to combine everything together naively.
    \end{proof}

\section{Penalty Term}\label{pena sec}
In this section, we present our results related to the penalty term.
\begin{Lem}\label{4x}
    If $r\sim\mathcal{F}_2$, then for all $x\ge 1$, we have
    \[
    \E[\,r\mid r\ge x]\leq 4x.
    \]
    \begin{proof}
        By the definition, we have
        \[
        \E[\,r\mid r\ge x]=\frac{\int_x^{+\infty} uf(u)\d u}{1-F(x)}\leq 2x^2 \int_x^{+\infty} uf(u)\d u\leq 2x^2\int_x^{+\infty} 2u^{-2}\d u=4x,
        \]
        where the first inequality used that $1-\mathrm{e}^{-x}\ge x/2$ when $0\leq x\leq 1$ and the second inequality used that $\e^{-1/u^2}\leq 1.$ 
    \end{proof}
\end{Lem}

\begin{Lem}\label{5md}
    Consider $r_1,\cdots,r_d\iidsim\mathcal{F}_2$ with $r_{(k)}$ as the $k$th order statistic for all $1\leq k\leq d$. Then for all $m\leq d$, the expectation of the largest $m$ numbers, say $\E\left[\sum_{k=d-m+1}^d r_{(k)}\right]$, is less than $5\sqrt{md}.$
    \begin{proof}
        Clearly, 
        \[
        \E\left[\sum_{k=d-m+1}^d r_{(k)}\right]\leq \sqrt{md}+ \E\left[\sum_{k=d-m+1}^d r_{(k)}\cdot\1_{\left\{r_{(k)}\ge\sqrt{d/m}\right\}}\right]\leq \sqrt{md}+\sum_{k=1}^d\E\left[ r_{k}\cdot\1_{\left\{r_{k}\ge\sqrt{d/m}\right\}}\right]
        .\]
        Then it suffices to note that, by Lemma \ref{4x}, we have
        \[
        \E\left[ r_{k}\cdot\1_{\left\{r_{k}\ge\sqrt{d/m}\right\}}\right]\leq 4\sqrt{d/m}\P\left(r_{k}\ge\sqrt{d/m}\right)=4\sqrt{d/m}\left(1-\mathrm{e}^{-m/d}\right)\leq 4\sqrt{m/d},
        \]
        where the final inequality used that $1-\e^{-x}\leq x.$
    \end{proof}
\end{Lem}

\subsection{Proof for Lemma \ref{phistar}}\label{proof phistar}
\begin{proof}
        If $a=\nabla\Phi(\lambda)$, which implies that $a\in\pd\Phi(\lambda)$, then by Lemma \ref{xxstar}, we have
        \[
        \Phi^*(a)=\agp{\lambda,a}-\Phi(\lambda)=\E[\agp{\lambda,A}]-\Phi(\lambda)=\E[\agp{\lambda,A}]-\E[\agp{r+\lambda,A}]=-\E[\agp{r,A}],
        \]
        where we used that $\E[A]=\phi(-\lambda)=a$ in the second equality and the third equality follows from the definition of $\Phi$. If $a\in\mathcal{A}$, for all $x\in\R^d$, we have $\Phi(x)\ge \E[\agp{r+x,a}]$, which implies that 
        \[
        \Phi^*(a)\leq\sup_{x\in\R^d} \agp{x,a}-\E[\agp{r+x,a}]=-\E[\agp{r,a}].
        \]
    \end{proof}

\subsection{Proof for Lemma \ref{pen1}}\label{proof pen1}
\begin{proof}
        By Lemma \ref{phistar},
        \begin{equation}\label{phiaphialam}
           \Phi^*(a)-\Phi^*(\phi(\lambda))\leq\E[\agp{r,A-a}], 
        \end{equation}
        where $A=\argmax_{a\in\mathcal{A}}\agp{r+\lambda,a}$. Then for the first result, since $r\in\R^{+d}$, 
        \[
        \Phi^*(a)-\Phi^*(\phi(\lambda))\leq \E[\max_{a'\in\mathcal{A}}\agp{r,a'}],
        \]
        which is less than $5\sqrt{md}$ by Lemma \ref{5md}.
        
        \noindent
        For the second result, W.L.O.G., we assume that $\lambda_1\leq\cdots\leq\lambda_d$, then $a=(\underbrace{1,\cdots,1}_{m \text{ of } 1},\underbrace{0,\cdots,0}_{d-m \text{ of } 0})$ and by \eq{phiaphialam},
        \[
        \Phi^*(a)-\Phi^*(\phi(\lambda))\leq\sum_{i=m+1}^d\E[r_i\1_{\{A_i=1\}}].
        \]
        By the definition of $A$, for all $i>m$, we have
        \[
        \begin{aligned}
            \E[r_i\1_{\{A_i=1\}}]=&\E_{r_i}[r_i\E[\1_{\{A_i=1\}}\mid r_i=x+\lambda_i]]\\
            =&\int_{-\lambda_i}^{\infty} \frac{2(x+\lambda_i)}{(x+\lambda_i)^3} e^{-1/(x+\lambda_i)^2} \sum_{s=0}^{m-1} \sum_{\mathcal{I} \subseteq  \{1,\dots,d\} \setminus \{i\}, |\mathcal{\mathcal{I}}|=s} \left[ \prod_{q \in \mathcal{I}} (1 - F(x+\lambda_q)) \prod_{q \notin \mathcal{I}, q \neq i} F(x+\lambda_q) \right] \d x\\
            =&2V_{i,2}(\lambda)\leq 2\underline{\lambda}_i^{-1},
        \end{aligned}
        \]
        where we used Lemma \ref{monontone1} in the final inequality. This completes our proof.
    \end{proof}

\section{Probability of the Best Action
}\label{w_star}
In this section, we present our results related to the lower bound for the regret.
\begin{Lem}\label{wstarbound}
    For all $\lambda\in\R^d$, let
    \[
    w_\star=\P\{\min\limits_{1\leq i\leq m}\left(r_i-\lambda_i\right)\ge \max\limits_{m+1\leq i\leq d}\left(r_i-\lambda_i\right)\},
    \]
    where $r_1,\cdots,r_d\iidsim\mathcal{F}_2.$ Then, 
    %if $\lambda_1\leq\cdots\leq\lambda_d$, 
    we have the followings:
    \begin{enumerate}
        \item If $\sum\limits_{i=m+1}^d \underline{\lambda}_i^{-2}<\frac{1}{2m},$ then $
        %\frac12\leq 
        w_\star \ge \frac12
        %\leq 1-\frac{1}{4\mathrm{e}}\sum\limits_{i=m+1}^d\underline{\lambda}_i^{-2}
        $.
        \item If $\sum\limits_{i=m+1}^d \underline{\lambda}_i^{-2}\ge \frac{1}{2m},$ then $w_\star\leq 1-\frac{1}{16m}$.
    \end{enumerate} 
    \begin{proof}
        If $\sum\limits_{i=m+1}^d \underline{\lambda}_i^{-2}=+\infty$, then there exists $1\leq i\leq m<j\leq d$ such that $\lambda_i\ge\lambda_j$. W.L.O.G., we assume that $\lambda_m\ge \lambda_{m+1}$. Denote that $X_i=r_i-\lambda_i$ for all $1\leq i\leq d$, then clearly,
        \[
        \begin{aligned}
            w_\star=&\P\{\text{$X_1,\cdots,X_m$ are the $m$ largest values among $X_1,\cdots,X_d$}\}\\
            \leq &\P\{\text{$X_1,\cdots,X_{m-1},X_{m+1}$ are the $m$ largest values among $X_1,\cdots,X_d$}\}:=w'_{\star}.
        \end{aligned}
        \]
        Note that $w_\star+w'_\star\leq 1$, then we have $w_\star\leq 1/2\leq 1-\frac{1}{16m}.$

        Then we assume that $\sum\limits_{i=m+1}^d \underline{\lambda}_i^{-2}<+\infty$, which implies that $\max\limits_{1\leq i\leq m}\lambda_i< \max\limits_{m+1\leq i\leq d}\lambda_i$. W.L.O.G., we assume that  $\lambda_1\leq\cdots\leq\lambda_m<\lambda_{m+1}\leq\cdots\leq\lambda_d$. Hence, $\max\limits_{\mathcal{I},|\mathcal{I}|<m}\min\limits_{j\notin \mathcal{I}}\lambda_j=\lambda_m$ and $\underline{\lambda}_i=\lambda_i-\lambda_m$ for all $i>m$. 
        By the definition of $w_\star$, 
        \begin{equation}\label{wstar}
        w_{\star}=2\int_{-\lambda_{m+1}}^{+\infty} \sum_{i=m+1}^d (x+\lambda_i)^{-3} \mathrm{e}^{-\sum\limits_{j=m+1}^d(x+\lambda_j)^{-2}}\prod_{1\leq k\leq m}\left(
        1-\mathrm{e}^{-(x+\lambda_k)^{-2}}\1_{\{x+\lambda_k\ge 0\}}
        \right)\d x.
        \end{equation}
        We first prove the lower bound for $w_\star$. Since for all $k\leq m$, $\lambda_k\leq\lambda_{m}$, then
        \[
        w_{\star}\ge2\int_{-\lambda_{m+1}}^{+\infty} \sum_{i=m+1}^d (x+\lambda_i)^{-3} \mathrm{e}^{-\sum\limits_{j=m+1}^d(x+\lambda_j)^{-2}}\left(
        1-\mathrm{e}^{-(x+\lambda_m)^{-2}}\1_{\{x+\lambda_m\ge 0\}}
        \right)^m\d x.
        \]
        Then by Bernoulli's inequality, we have
        \[
        \begin{aligned}
            w_\star\ge& 2\int_{-\lambda_{m+1}}^{+\infty} \sum_{i=m+1}^d (x+\lambda_i)^{-3} \mathrm{e}^{-\sum\limits_{j=m+1}^d(x+\lambda_j)^{-2}}\left(
        1-m\mathrm{e}^{-(x+\lambda_m)^{-2}}\1_{\{x+\lambda_m\ge 0\}}
        \right)\d x\\
        =&2\int_{-\lambda_{m+1}}^{+\infty} \sum_{i=m+1}^d (x+\lambda_i)^{-3} \mathrm{e}^{-\sum\limits_{j=m+1}^d(x+\lambda_j)^{-2}}\d x-2m\int_{-\lambda_{m}}^{+\infty} \sum_{i=m+1}^d (x+\lambda_i)^{-3} \mathrm{e}^{-\sum\limits_{j=m+1}^d(x+\lambda_j)^{-2}-(x+\lambda_m)^{-2}}\d x\\
        =&1-m\left(
        1-\int_{-\lambda_{m}}^{+\infty} 2(x+\lambda_m)^{-3} \mathrm{e}^{-\sum\limits_{j=m+1}^d(x+\lambda_j)^{-2}-(x+\lambda_m)^{-2}}\d x
        \right)\\
        =&1-m\left(1-\int_{0}^{+\infty} 2z^{-3} \mathrm{e}^{-\sum\limits_{j=m+1}^d(z+\underline{\lambda}_j)^{-2}-z^{-2}}\d z\right)\\
        \ge&1-m\left(1-\mathrm{e}^{-\sum\limits_{j=m+1}^d\underline{\lambda}_j^{-2}}\int_{0}^{+\infty} 2z^{-3} \mathrm{e}^{-z^{-2}}\d z\right)\\
        =&1-m\left(1-\mathrm{e}^{-\sum\limits_{j=m+1}^d\underline{\lambda}_j^{-2}}\right),
        \end{aligned}
        \]
        where in the third line we used that $\d\left(\mathrm{e}^{-\sum\limits_{j=m+1}^d(x+\lambda_j)^{-2}}\right)=2\sum\limits_{i=m+1}^d (x+\lambda_i)^{-3}\mathrm{e}^{-\sum\limits_{j=m+1}^d(x+\lambda_j)^{-2}}$ and in the fourth line we used that $\underline{\lambda}_i=\lambda_i-\lambda_m$ for all $i>m$. Hence, when $\sum\limits_{i=m+1}^d \underline{\lambda}_i^{-2}<\frac{1}{2m}$, by that $-1+\mathrm{e}^{-x}\ge x$, we have
        \[
        w_\star\ge 1-\frac{m}{2m}=\frac12.
        \]

        We then show the upper bounds. By \eq{wstar}, clearly,
        \begin{equation}\label{lower}
                   \begin{aligned}
            w_\star\leq&2\int_{-\lambda_{m+1}}^{+\infty} \sum_{i=m+1}^d (x+\lambda_i)^{-3} \mathrm{e}^{-\sum\limits_{j=m+1}^d(x+\lambda_j)^{-2}}\left(
        1-\mathrm{e}^{-(x+\lambda_m)^{-2}}\1_{\{x+\lambda_m\ge 0\}}
        \right)\d x\\
        =&1-2\int_{-\lambda_{m}}^{+\infty} \sum_{i=m+1}^d (x+\lambda_i)^{-3} \mathrm{e}^{-\sum\limits_{j=m+1}^d(x+\lambda_j)^{-2}-(x+\lambda_m)^{-2}}\d x\\
        =&1-\int_{0}^{+\infty} 2\sum_{i=m+1}^d (z+\underline{\lambda}_i)^{-3} \mathrm{e}^{-\sum\limits_{j=m+1}^d(z+\underline{\lambda}_j)^{-2}-z^{-2}}\d z\\
        \leq &1-\int_{\underline{\lambda}_{m+1}}^{+\infty} 2\sum_{i=m+1}^d (z+\underline{\lambda}_i)^{-3} \mathrm{e}^{-\sum\limits_{j=m+1}^d(z+\underline{\lambda}_j)^{-2}-z^{-2}}\d z.
        \end{aligned}
        \end{equation}
        \noindent If $\sum\limits_{i=m+1}^d \underline{\lambda}_i^{-2}\ge \frac{1}{2m}$, note that when $z\ge \underline{\lambda}_{m+1}$, 
        $
        \frac{1}{z^2}\leq \frac{4}{(z+\underline{\lambda}_{m+1})^2},
        $ then
        \[
        \sum\limits_{j=m+1}^d(z+\underline{\lambda}_j)^{-2}+z^{-2}\leq5\sum\limits_{j=m+1}^d(z+\underline{\lambda}_j)^{-2}.
        \]
        Hence, by \eq{lower},
        \[
        \begin{aligned}
            w_\star\leq& 1-\int_{\underline{\lambda}_{m+1}}^{+\infty} 2\sum_{i=m+1}^d (z+\underline{\lambda}_i)^{-3} \mathrm{e}^{-5\sum\limits_{j=m+1}^d(z+\underline{\lambda}_j)^{-2}}\d z\\
            = &1-\frac15\left(
            1-\mathrm{e}^{-5\sum\limits_{j=m+1}^d(\underline{\lambda}_{m+1}+\underline{\lambda}_j)^{-2}}
            \right)\\
            \leq& 1-\frac15\left(
            1-\mathrm{e}^{-\frac54\sum\limits_{j=m+1}^d\underline{\lambda}_j^{-2}}
            \right)\\
            \leq &1-\frac15\left(
            1-\mathrm{e}^{-\frac{5}{8m}}
            \right)\leq 1-\frac{1}{16m},
        \end{aligned}
        \]
        where we used that $1-\mathrm{e}^{-x}\ge \frac{x}2$ when $x\in[0,1]$ in the final inequality.
     \end{proof}
\end{Lem}

\begin{Lem}\label{phiilower}
    If $\sum\limits_{i=m+1}^d \underline{\lambda}_i^{-2}<\frac{1}{2m},$ then for all $m< i\leq d$, $\phi_i(\lambda)\ge\frac{1}{4\mathrm{e}}\underline{\lambda}_i^{-2}.$
    \begin{proof}
        By the definition of $\phi_i$ in \eq{defi phi}, clearly, for all $m< i\leq d$,
    \begin{equation}\label{imd}
        \begin{aligned}
            \phi_i(\lambda)\ge&\P\{\text{
    $r_i-\lambda_i$ is the largest in $r_{m}-\lambda_{m},\cdots,r_d-\lambda_d$
    }\}\\
    =&\int_{-\min\limits_{m\leq j\leq d} \lambda_j}^{\infty} \frac{2}{\left(z+\lambda_i\right)^3} \exp \left(-\sum\limits_{j=m}^d \frac{1}{\left(z+\lambda_{j}\right)^2}\right) \mathrm{d} z.
        \end{aligned}
    \end{equation}
    Since $\sum\limits_{i=m}^d \underline{\lambda}_i^{-2}< \frac{1}{2m}$, similar to Lemma \ref{wstarbound}, W.L.O.G., we assume that  $\lambda_1\leq\cdots\leq\lambda_m<\lambda_{m+1}\leq\cdots\leq\lambda_d$. Hence, $\max\limits_{\mathcal{I},|\mathcal{I}|<m}\min\limits_{j\notin \mathcal{I}}\lambda_j=\lambda_m$ and $\underline{\lambda}_i=\lambda_i-\lambda_m$ for all $i>m$. Therefore, by \eq{imd}, we have
    \[
    \begin{aligned}
        \phi_i(\lambda)\ge&\int_{0}^{\infty} \frac{2}{\left(z+\underline{\lambda}_i\right)^3} \exp \left(-\sum\limits_{j=m}^d \frac{1}{\left(z+\underline{\lambda}_{j}\right)^2}\right) \mathrm{d} z\\
        \ge&\int_{\underline{\lambda}_{m+1}}^{\infty} \frac{2}{\left(z+\underline{\lambda}_i\right)^3} \exp \left(-\sum\limits_{j=m}^d \frac{1}{\left(z+\underline{\lambda}_{j}\right)^2}\right) \mathrm{d} z.
    \end{aligned}
    \]
    Note that when $z\ge \underline{\lambda}_{m+1}$, 
        \[
        \sum\limits_{j=m}^d \frac{1}{\left(z+\underline{\lambda}_{j}\right)^2}=\sum\limits_{j=m+1}^d(z+\underline{\lambda}_j)^{-2}+z^{-2}\leq \sum\limits_{j=m+1}^d\underline{\lambda}_j^{-2}+\underline{\lambda}_{m+1}^{-2}\leq 2\sum\limits_{j=m+1}^d\underline{\lambda}_j^{-2}<\frac{1}{m}\leq 1,
        \]
        then by \eq{lower},
        % \[
        % w_\star\leq1-\mathrm{e}^{-1}\int_{\underline{\lambda}_{m+1}}^{+\infty} 2\sum_{i=m+1}^d (z+\underline{\lambda}_i)^{-3} \d z=1-\mathrm{e}^{-1}\sum_{i=m+1}^d\left(\underline{\lambda}_i+\underline{\lambda}_{m+1}\right)^{-2}\leq 1-\frac{1}{4\mathrm{e}}\sum\limits_{i=m+1}^d\underline{\lambda}_i^{-2},
        % \]
        \[\phi_i(\lambda)\ge\mathrm{e}^{-1}\int_{\underline{\lambda}_{m+1}}^{+\infty} 2 (z+\underline{\lambda}_i)^{-3} \d z=\mathrm{e}^{-1}\left(\underline{\lambda}_i+\underline{\lambda}_{m+1}\right)^{-2}\ge \frac{1}{4\mathrm{e}}\underline{\lambda}_i^{-2},
        \]
        where in the final inequality we used that $\underline{\lambda}_{m+1}\leq \underline{\lambda}_i$ for all $i\ge m+1$.
    \end{proof}
\end{Lem}

\begin{Lem}\label{wstarwti}
    Use the definition of $w_{\star}$ in Lemma \ref{wstarbound}, then we have
    \[
    1-w_{\star}\leq\sum_{i=m+1}^d \phi_i(\lambda).
    \]
    \begin{proof}
        Clearly,
        \[
        \begin{aligned}
            1-w_{\star}=&\P\{\min\limits_{1\leq i\leq m}\left(r_i-\lambda_i\right)< \max\limits_{m+1\leq i\leq d}\left(r_i-\lambda_i\right)\}\\
            =&\P\left\{
            \bigcup_{i=m+1}^d\left\{
                \text{$r_i-\lambda_i$ is among the top $m$ largest values in $r_1-\lambda_1,\cdots,r_d-\lambda_d$
    }
            \right\}
            \right\}\\
            \leq&\sum_{i=m+1}^d\phi_i(\lambda).
        \end{aligned}
        \]
    \end{proof}
\end{Lem}

\section{Auxiliary Lemma}\label{auxi sec}
\begin{Lem}[Theorem 23.5 in \cite{ca}]\label{xxstar}
    For any continuous convex function $g$ and any vector $x$, the following conditions on a vector $x^*$ are equivalent to each other:
    \begin{enumerate}
        \item $x^*\in\partial g(x)$.
        \item $x\in\partial g^*(x^*)$.
        \item $g(x)+g^*(x^*)=\agp{x,x^*}.$
    \end{enumerate}
\end{Lem}
\begin{Lem}[Generalized Pythagoras Identity]\label{xyz}
    \[
    D_\Phi(x,y)+D_\Phi(z,x)-D_\Phi(z,y)=\agp{\nabla\Phi(x)-\nabla\Phi(y),x-z}.
    \]
    \begin{proof}
        It suffices to expand the left hand by the definition that
        \[
        D_\Phi(x,y)=\Phi(x)-\Phi(y)-\agp{x-y,\nabla\Phi(y)}.
        \]
    \end{proof}
\end{Lem}
\begin{Lem}\label{uv xy}
    If $u=\nabla\Phi(x)$ and $v=\nabla\Phi(y)$, then
    \[
    D_{\Phi}(y,x)=\Phi^*(u)-\Phi^*(v)-\agp{u-v,y}.
    \]
    \begin{Rema}
        Informally, this is just the folklore that $D_{\Phi}(y,x)=D_{\Phi^*}(u,v)$. However, it remains unproven whether $\Phi^*$ is differentiable everywhere, making it inconvenient to discuss its Bregman divergence directly.
    \end{Rema}
    \begin{proof}
        By Lemma \ref{xxstar}, we have
        \[
        \Phi^*(u)=\agp{u,x}-\Phi(x), \Phi^*(v)=\agp{v,y}-\Phi(y).
        \]
        Then the right hand equals
        \[
        \Phi(y)-\Phi(x)-\agp{u,y-x}=\Phi(y)-\Phi(x)-\agp{\nabla\Phi(x),y-x}=D_{\Phi}(y,x).
        \]
    \end{proof}
\end{Lem}
\begin{Lem}\label{k2ge}
    If $K$ is sampled from the Geometric distribution with parameter $p\in(0,1)$, then $\E[K^2]\leq \frac{2}{p^2}.$ Furthermore, for all $n\in\N^+$, $\E[K-K\wedge n]=p^{-1}(1-p)^n$.
    \begin{proof}
    For the first result,
        \[
        \E[K^2]=\E[K]^2+\operatorname{Var}(K)=\frac{1}{p^2}+\frac{1-p}{p^2}\leq\frac{2}{p^2}.
        \]
    For the second result, by direct calculation, we have
    \[
    \E[K-K\wedge n]=\sum_{k=n+1}^{+\infty}\P(K\ge k)=\sum_{k=n+1}^{+\infty}(1-p)^{k-1}=p^{-1}(1-p)^n.
    \]
    \end{proof}
\end{Lem}

\begin{Lem}\label{sumof}
    For all $n\in\N^+$,
    \[
    \sum_{k=1}^n\frac{1}{\sqrt{k}}\leq 2\sqrt{n}.
    \]
    \begin{proof}
        \[
        \sum_{k=1}^n\frac{1}{\sqrt{k}}\leq 1+\int_1^{n}\frac{1}{\sqrt{x}}\d x\leq 2\sqrt{n}.
        \]
    \end{proof}
\end{Lem}

% \begin{Lem}[Integral Chebyshev inequality]\label{ICI}
%     If $f, g:[a, b] \rightarrow \R$ are two monotonic functions of the same monotonicity, then
%     \[
%     \frac{1}{b-a} \int_a^b f(x) g(x) d x \geq\left[\frac{1}{b-a} \int_a^b f(x) d x\right]\left[\frac{1}{b-a} \int_a^b g(x) d x\right].
%     \]
% \end{Lem}

% \begin{Lem}\label{gammaxx}
%     For all $x>0$, we have $\Gamma(x+1)\leq 2x^x.$
%     \begin{proof}
%     Clearly,
%         \[
%         \Gamma(x+1)=\int_0^{\infty} z^x e^{-z} \d z 
% =\int_0^{\infty}\left(z^x e^{\frac{-z}{2}}\right) e^{\frac{-z}{2}} d z 
% \leq (2 x)^x e^{-x} \int_0^{\infty} e^{\frac{-z}{2}} \d z
% =2(2 x)^x e^{-x}\leq 2x^x,
%         \]
%     where we used that $z^x e^{-z/2}\leq(2x)^x e^{-x}$ for all $z\ge 0$ and $e<2.$
%     \end{proof}
% \end{Lem}

\begin{Lem}\label{2etat}
    For all $x\in(0,1]$, let $f(x)=\sum\limits_{t=1}^{+\infty} 2^{-\sqrt{t}x}$, then there exists $C>0$ such that $f(x)\leq \frac{C}{x^2}.$
    \begin{proof}
        It suffices to note that
        \[
        f(x)\leq\int_0^{+\infty} 2^{-\sqrt{t}x}\d t=2\int_0^{+\infty}u2^{-ux} \d u=\frac{2}{x^2}\int_0^{+\infty}v2^{-v}\d v.
        \]
    \end{proof}
\end{Lem}

\begin{Lem}\label{ratio1-f}
    For all $\mu\in\R$ and $y\ge x>0$, if $y\ge 1$, then we have
    \[
    \frac{1-F(x+\mu)}{1-F(y+\mu)}\leq 8y^2.
    \]
    \begin{proof}
        If $x+\mu\leq y$, then
        \[
        \frac{1-F(x+\mu)}{1-F(y+\mu)}\leq\frac{1}{1-F(y+\mu)}.
        \]
        When $y+\mu\leq 0$, the right hand equals $1$ and is clearly less than $8y^2$. Otherwise, $y+\mu>0$ and then it suffices to note that
        \[
        1-F(y+\mu)=1-e^{-(y+\mu)^{-2}}\ge 1-e^{-y^{-2}/4}\ge y^{-2}/8,
        \]
        where we used $y+\mu\leq y+x+\mu\leq 2y$ in the first inequality and $1-e^{-x}\ge x/2$ when $0\leq x\leq 1$ in the second inequality.

        If $x+\mu>y$, then $y+\mu\ge x+\mu>y\ge 1$ and similarly,
        \[
        1-F(y+\mu)=1-e^{-(y+\mu)^{-2}}\ge(y+\mu)^{-2}/2.
        \]
        Also, since for all $x\ge 0$, $1-e^{-x}\leq x$, we then have
        \[
        1-F(x+\mu)=1-e^{-(x+\mu)^{-2}}\leq(x+\mu)^{-2}.
        \]
        Therefore,
        \[
        \frac{1-F(x+\mu)}{1-F(y+\mu)}\leq 2\left(\frac{y+\mu}{x+\mu}\right)^2<2\left(\frac{y+y-x}{y}\right)^2\leq 8\leq 8y^2.
        \]
    \end{proof}
\end{Lem}

\begin{Lem}[Theorem 1.2.6 in \cite{durrett2019probability}]\label{gautail}
    For $x > 0$,
    \[
    \left(x^{-1} - x^{-3}\right) \exp(-x^2/2) 
\leq \int_x^{\infty} \exp(-y^2/2) \d y 
\leq x^{-1} \exp(-x^2/2).
    \]
\end{Lem}

\begin{Lem}\label{moment}
    If $Y\sim\N(0,\sigma^2)$ and $\mu\ge\sigma>0$, then for all $k\in\mathcal{N}$, we have
    \[
    \E[Y^k\1_{\{Y\ge \mu\}}]\leq k!!\cdot\sigma\mu^{k-1}\mathrm{e}^{-\frac{\mu^2}{2\sigma^2}}/\sqrt{2\pi},
    \]
    where we define that $0!!=1$.
    \begin{proof}
        If $k=0$, by Lemma \ref{gautail}, we have
        \[
        \E[\1_{\{Y\ge \mu\}}]=\E[\1_{\{\frac{Y}{\sigma}\ge\frac{\mu}{\sigma}\}}]\leq \frac{\sigma}{\mu}e^{-\frac{\mu^2}{2\sigma^2}}/\sqrt{2\pi}.
        \]
        If $k=1$, 
        \[
        \E[Y\1_{\{Y\ge \mu\}}]=\frac{1}{\sqrt{2\pi\sigma^2}}\int_{\mu}^{+\infty} y \mathrm{e}^{-\frac{y^2}{2\sigma^2}}\d y=\frac{1}{\sqrt{2\pi\sigma^2}}\sigma^2e^{-\frac{\mu^2}{2\sigma^2}}.
        \]
        Assume that the statement holds for all integers $0,1, \ldots, k-1$, then
        \[
        \begin{aligned}
            \E[Y^k\1_{\{Y\ge \mu\}}]=&\frac{1}{\sqrt{2\pi\sigma^2}}\int_{\mu}^{+\infty} y^k \mathrm{e}^{-\frac{y^2}{2\sigma^2}}\d y=\frac{\sigma^2}{\sqrt{2\pi\sigma^2}}\int_{\mu}^{+\infty} y^{k-1} \d\left(-\mathrm{e}^{-\frac{y^2}{2\sigma^2}}\right)\\
            =&\frac{\sigma^2}{\sqrt{2\pi\sigma^2}}\mu^{k-1}\mathrm{e}^{-\frac{\mu^2}{2\sigma^2}}+(k-1)\sigma^2\E[Y^{k-2}\1_{\{Y\ge \mu\}}] .
        \end{aligned}
        \]
        Then it suffices to note that $\sigma\leq\mu$ and $1+(k-1)\cdot (k-2)!!\leq k\cdot (k-2)!!=k!!.$
    \end{proof}
\end{Lem}
\begin{Lem}\label{2mu}
    For any random variable \( X \), let its density be denoted by \( f(x) \). If there exist \(\mu\ge2\sigma > 0\) and $\mu'>\mu$ such that
\[
\frac{\d \log f}{\d x}(x) \leq -x/\sigma^2, \quad \forall \mu'\ge x \ge \mu,
\]
then for all $k\in\mathcal{N}^+$, if $\mu'\ge2\mu$, we have
\[
\E[X^k \, \1_{\{\mu'\ge X \ge \mu\}}]\leq 2k!!\cdot\mu^k.
\]
\begin{proof}
    %W.L.O.G, one can assume $\mu'\ge 2\mu$. 
    Consider $Y\sim\N(0,\sigma^2)$, then we will show that 
    \begin{equation}\label{aim}
        \E[X^k \, \1_{\{\mu'\ge X \ge \mu\}}]\leq \E[Y^k\,|\,\mu'\ge Y\ge\mu],
    \end{equation}
    which just equals
    \begin{equation}\label{cony}
        \frac{\E[Y^k\1_{\{\mu'\ge Y\ge \mu\}}]}{\P(\mu'\ge Y\ge \mu)}\leq \frac{\E[Y^k\1_{\{Y\ge \mu\}}]}{\P(\mu'\ge Y\ge \mu)}=\frac{\sqrt{2\pi}\E[Y^k\1_{\{Y\ge \mu\}}]}{\int_{\frac{\mu}{\sigma}}^{\frac{\mu'}{\sigma}}e^{-\frac{x^2}{2}}\d x}\leq \frac{k!!\cdot\sigma\mu^{k-1}\mathrm{e}^{-\frac{\mu^2}{2\sigma^2}}}{\int_{\frac{\mu}{\sigma}}^{\frac{\mu'}{\sigma}}e^{-\frac{x^2}{2}}\d x}
    % \leq \frac{\sigma e^{-\frac{\mu^2}{2\sigma^2}}}{\left(\frac{\sigma}{\mu}-\frac{\sigma^3}{\mu^3}\right)e^{-\frac{\mu^2}{2\sigma^2}}}=\frac{\mu}{1-\frac{\sigma^2}{\mu^2}}
    ,
    \end{equation}
    where we applied Lemma \ref{moment} in the last inequality. By Lemma \ref{gautail}, we have
    \begin{equation}\label{mumu'}
        \int_{\frac{\mu}{\sigma}}^{\frac{\mu'}{\sigma}}e^{-\frac{x^2}{2}}\d x\ge \left(\frac{\sigma}{\mu}-\frac{\sigma^3}{\mu^3}\right)e^{-\frac{\mu^2}{2\sigma^2}}-\frac{\sigma}{\mu'}e^{-\frac{\mu'^2}{2\sigma^2}}\ge\left(\frac{\sigma}{\mu}-\frac{\sigma^3}{\mu^3}\right)e^{-\frac{\mu^2}{2\sigma^2}}-\frac{\sigma}{2\mu}e^{-\frac{2\mu^2}{\sigma^2}},
    \end{equation}
    where we used $\mu'\ge 2\mu$ in the last inequality. Since $\mu\ge \sigma$, we have $e^{-\frac{2\mu^2}{\sigma^2}}\leq \frac{1}{2}e^{-\frac{\mu^2}{2\sigma^2}}$. Hence, combining \eq{cony} and \eq{mumu'}, we have
    \[
    \E[Y^k\,|\,\mu'\ge Y\ge\mu]\leq \frac{k!!\cdot\sigma\mu^{k-1}\mathrm{e}^{-\frac{\mu^2}{2\sigma^2}}}{\left(\frac{3\sigma}{4\mu}-\frac{\sigma^3}{\mu^3}\right)e^{-\frac{\mu^2}{2\sigma^2}}}\leq \frac{2k!!\cdot\mu^k}{\frac{3}{2}-\frac{2\sigma^2}{\mu^2}}.
    \]
    The right hand is clearly less than $2\mu^k$ when $\mu\ge 2\sigma.$ To show \eq{aim}, it suffices to show that for all $0\leq t\leq \mu'$, 
    \begin{equation}\label{tailxy}
        \P(X \, \1_{\{\mu'\ge X \ge \mu\}}\ge t)\leq \P(Y\ge t|\,\mu'\ge Y\ge\mu),
    \end{equation}
    which holds when $t\leq \mu$ because the right hand becomes $1$ by the definition. In fact, we can prove a stronger result: for all $\mu'\ge t>\mu$,
    \begin{equation}\label{aimratio}
        \frac{\P(X \, \1_{\{\mu'\ge X \ge \mu\}}\ge t)}{\P(X \, \1_{\{\mu'\ge X \ge \mu\}}\ge \mu)}\leq \frac{\P(Y\ge t|\,\mu'\ge Y\ge\mu)}{\P(Y\ge \mu|\,\mu'\ge Y\ge\mu)}.
    \end{equation}
    If there exists \( t_0 > \mu \) that violates \eq{aimratio}, then we have $\frac{\P(X \, \1_{\{\mu'\ge X \ge \mu\}}\ge t_0)}{\P(X \, \1_{\{\mu'\ge X \ge \mu\}}\ge \mu)}> \frac{\P(Y\ge t_0|\,\mu'\ge Y\ge\mu)}{\P(Y\ge \mu|\,\mu'\ge Y\ge\mu)}$, which is equivalent to 
    \[
        \frac{\P(X \, \1_{\{\mu'\ge X \ge \mu\}}\ge t_0)}{\P(\mu\leq X \, \1_{\{\mu'\ge X \ge \mu\}}< t_0)}> \frac{\P(Y\ge t_0|\,\mu'\ge Y\ge\mu)}{\P(Y<t_0|\,\mu'\ge Y\ge\mu)},
    \]
    because $\frac{x}{x+1}$ increases with positive $x$. Then we have
    \begin{equation}\label{aimint}
        \frac{\int_{t_0}^{\mu'}f(x)\d x}{\int_{\mu}^{t_0}f(x)\d x}>\frac{\int_{t_0}^{\mu'}e^{-\frac{x^2}{2\sigma^2}}\d x}{\int_{\mu}^{t_0}e^{-\frac{x^2}{2\sigma^2}}\d x}. 
    \end{equation}
    By homogeneity, we may assume $f(t_0)=e^{-\frac{t_0^2}{2\sigma^2}}$. Now since for all $\mu'\ge x\ge \mu$, we have $\frac{\d \log f(x)}{\d x} \leq -x/\sigma^2,$ then for all $x\ge t_0$, $f(x)\leq e^{-\frac{x^2}{2\sigma^2}}$; for all $\mu\leq x< t_0$, $f(x)\ge e^{-\frac{x^2}{2\sigma^2}}$, which contradicts with \eq{aimint}. 
\end{proof}

\end{Lem}

%%%%%%%%%%%%%%%%%%%%%%%%%%%%%%%%%%%%%%%%%%%%%%%%%%%%%%%%%%%%

\end{document}